\documentclass{article}

\usepackage{PRIMEarxiv}

\usepackage[utf8]{inputenc} 
\usepackage[T1]{fontenc}    
\usepackage{hyperref}       
\usepackage{url}            
\usepackage{booktabs}       
\usepackage{amsfonts}       
\usepackage{nicefrac}       
\usepackage{microtype}      
\usepackage{lipsum}
\usepackage{fancyhdr}       
\usepackage{graphicx}       
\graphicspath{{media/}}     
\usepackage{amsmath}
\usepackage{amssymb}
\usepackage{makecell}   
\usepackage{booktabs}

\title{Towards An LLM-Driven Unified Conversion Framework for BT and FSM in Autonomous Intelligent Systems
\thanks{\textit{\underline{Citation}}: 
\textbf{Authors. Title. Pages.... DOI:000000/11111.}} 
}

\author{
  Zhang Qi, Yang Shuo, Zhu Zhengqiu, Zhou Peng, Jiao Peng\\
  College of Systems Engineering, National University of Defense Technology \\
  CHANGSHA, HUNAN, China\\
  \texttt{zhangqiy123@nudt.edu.cn} \\
}

\begin{document}
\maketitle

\begin{abstract}
Finite state machine (FSM) and behavior trees (BT) are widely adopted behavioral modeling paradigms for autonomous intelligent systems. While functionally equivalent and inter-convertible in principle, existing transformation methods between FSM and BT face major challenges in preserving behavioral completeness and avoiding model complexity explosion. To overcome these issues, we propose an LLM-driven unified conversion framework  that enables automatic, efficient, and semantically consistent transformation between FSM and BT. Specifically, a novel loop execution BT structure is designed for LLM to accurately capture the loop structure in FSM, thereby preserving behavioral completeness. To mitigate the state explosion problem in BT-to-FSM conversion, a depth compression strategy is introduced with LLM prompt to eliminate redundant control nodes, complemented by differentiated hierarchical conversion rules that collectively reduce the number of required sub-FSM. Simulation experiments in multiple autonomous decision-making scenarios demonstrate that the proposed framework enables an accurate and automated bidirectional conversion between FSM and BT. Furthermore, it significantly enhances the scalability and maintainability of generated models compared to traditional approaches, providing a practical solution for behavior model conversion in consumer-grade autonomous intelligent systems such as service robots, game agents, and smart home devices
\end{abstract}


\section{Introduction}
In consumer autonomous intelligent (AISs) applications such as home and service robots [1], smart home appliances, interactive game agents [2], and smart electric vehicles [3], behavioral modeling is a key component in achieving a closed-loop of autonomous perception, decision-making, and execution. Since these consumer electronics systems are typically deployed in dynamic environments and interact directly with end users, their behavioral logic must be interpretable, maintainable, verifiable, and easily portable across software architectures and hardware platforms. FSM [4-5] are widely used in such systems due to their advantages, including explicit state transitions, low runtime overhead, and verifiable execution paths. These characteristics are particularly important for consumer electronics, as systems typically require predictable behavior, convenient debugging capabilities, and safety-related traceability during product development, testing, and updates. However, their single architecture faces a state-explosion problem in dynamic environments[6], when there are \textit{n} states, the number of state transitions is \textit{O(n²)}. This compromises explainability and safety compliance due to the inability to audit decision logic.

In contrast, behavior tree (BT) [6-8] demonstrate superior flexibility and maintainability in complex decision-making scenarios through modular sub-tree reuse. This enables independent development, testing, and deployment, high responsiveness in handling task interruptions, as well as hierarchical task decomposition. However, the introduction of unexplained polling latency and opaque sub-tree interactions is a fundamental barrier for applications that require human-in-the-loop transparency. These paradigms exhibit significant complementarity across dimensions, including real-time operation versus flexibility and determinism versus extensibility [9-10].

FSM remain widely used in robotic and gaming systems for their explicit state transitions and ease of verification, while BT are increasingly adopted in autonomous intelligent systems for their modularity and reactivity [6,11]. Therefore, seamless bidirectional FSM-BT conversion is essential for technology migration, hybrid system design, and long-term maintenance, especially in consumer electronics where behavioral logic frequently evolves through software iteration, firmware updates, platform migration, and behavioral logic refactoring. The conversion of FSM-to-BT improves the modularity and reactivity of legacy control modules, whereas the conversion of BT-to-FSM supports formal verification, debugging, runtime monitoring, certification and integration with existing FSM-based infrastructures [12-14]. However, manual conversion is time-consuming and prone to structural inconsistency, semantic deviation, and behavioral misalignment caused by the mismatch between event-driven FSM and poll-based BT. Thus, an automated and semantically consistent conversion framework is needed to reduce engineering costs and improve maintainability.

Conventional conversion methods[15-16] typically rely on fixed templates and mapping rules built on expert knowledge; therefore, they require the input model to adopt a predefined formal representation and satisfy specific structural assumptions. When the input is presented in non-prescribed formats—such as images, text, or structured tables—or employs new modeling conventions and transformation semantics, traditional methods often require additional format conversion and model preprocessing, or even manual extension or rewriting of transformation rules. This reliance on specific input representations and manually defined rules limits the generalizability of input representations and increases the cost of system adaptation. Furthermore, target models generated by traditional methods are prone to structural redundancy, resulting in low modularity and readability, which in turn increases the difficulty of subsequent feature expansion, debugging, and maintenance.

Recent advances in large language models (LLMs) [17-19] open new pathways for the conversion of intelligent behavioral models. With enhanced natural language understanding and logical reasoning capabilities, LLMs can parse control logic, preserve semantic equivalence, and verify behavioral consistency during the FSM–BT transformation. This substantially reduces the development threshold and improves cross-paradigm alignment accuracy. LLMs[20-22] conversion methods face three primary challenges when applied to complex scenarios: (1) LLMs' insufficient comprehension of intricate behavioral logic may result in semantic information loss; (2) the structural mapping between FSM and BT remains unresolved, leading to incomplete semantic conversion in certain instances; (3) Inconsistencies in conversion results, as LLMs may generate multiple viable conversion schemes. These challenges render current LLM-based approaches insufficient to meet the practical demands of efficient and reliable conversion between behavioral modeling paradigms.

To overcome these limitations, this paper proposes an LLM-driven unified conversion framework (LLMDUCF) for the bidirectional transformation between FSM and BT. LLMDUCF treats the LLM as a high-level behavioral model conversion engineer, whose reasoning process is constrained by domain-specific conversion rules, structural mappings, and task-oriented prompts. Under these constraints, the LLM parses the source behavioral model, reasons about its execution semantics, and synthesizes a structurally correct target model, thereby enabling automated and semantically consistent model conversion.

In summary, the key contributions of this paper are as follows. 

1) We propose LLMDUCF, an LLM-driven unified framework that formulates bidirectional FSM-BT conversion as constrained semantic reasoning, enabling LLMs to perform structural recognition, semantic decomposition, execution-logic reasoning, target-model synthesis, and structural correctness under domain-specific conversion constraints.

2) We propose a loop execution behavior tree (LEBT) for behavior-preserving FSM-to-BT conversion. The LEBT represents typical FSM control-flow patterns, including entry, loop, chain, and branch structures; guided by the proposed mapping specifications, the LLM recognizes these structures, reasons about FSM transition semantics, and synthesizes a complete BT that preserves loop behavior and maintains consistent execution logic.

3) We propose a BT depth-compression strategy and a differentiated hierarchical conversion mechanism for scalable BT-to-FSM conversion. These mechanisms guide the LLM to merge redundant homogeneous control nodes, decompose compressed BTs into semantic conversion units, infer the execution logic of sequence and selector nodes, and generate the corresponding FSM structures, thereby reducing BT depth, mitigating sub-FSM explosion, and improving the maintainability of the generated FSM.

The remainder of this article is organized as follows: Section II briefly discusses related studies. Section III presents the proposed bidirectional FSM and BT conversion framework for AISs. The simulation results are presented in Section IV, Section V concludes the article.

\section{BACKGROUND AND RELATED WORKS}
\label{sec:headings}

\subsection{Conventional Model Conversion Approaches}
Colledanchise et al. [15] have proposed explicit state-variable mapping by converting each FSM state to a BT node and transitioning to global variable-driven selector/sequence combinations to achieve semantic equivalence. Their state simulation method interprets BT execution states (running/success/failure) as FSM transition triggers, thereby enabling BT-to-FSM conversion through chained state transitions. Marzinotto et al. [16] have established a unified BT framework that maps FSM transitions to condition nodes and actions to action nodes for automated FSM-to-BT conversion. For reverse conversion, they have designed recursive embedding rules to translate BTs sequence, selection, and root nodes into FSM structures. Martín et al. [23] have integrated symbolic planning to map high-level BT structures to FSM planning states to generate executable sequences and reconstruct FSM logic as  combinations of BT nodes. Hallen et al. [24] have achieved bidirectional conversion via functional decoupling, where FSMs govern when to act and BTs determine how to act.

Although existing conversion methods show certain effectiveness, several critical limitations must be addressed to meet the demands of autonomous intelligent systems. First, the generated FSM exhibit O(\textit{n²}) state growth as BT depth increases, which not only undermines model maintainability but can also lead to state explosion and degraded conversion performance. Second, the resulting BT often suffers from limited expressiveness and structural over-flattening, reducing readability, and introducing semantic inconsistencies. Third, these methods primarily provide transformation rules or algorithmic concepts and do not directly include complete behavioral model parsing modules or engineered transformation implementation modules.

\subsection{LLM-based Automated Conversion Approaches}
Recent research has explored the generation of LLM-based automated behavioral models. Gan et al. [25] have developed ChatFSM, which is a multiagent framework that reads FSM structures from codebases. They have used retrieval-augmented generation for context and guiding LLMs to modify the FSM code based on natural language instructions. Lin et al. [26] have proposed systematic markdown prompt templates with task-oriented prompt patches and step-by-step instructions for LLMs to focus on key design constraints. This improves the accuracy of FSM generation  through chain-of-thought reasoning. Swick et al. [27] have proposed a state machine synthesis method based on LLMs to allow humans to plan robot movements. This method reduces the burden of robot programming while ensuring the safety and human-collaborative nature of conventional motion planning. Zhou et al. [28] have proposed an LLM–BT method that uses the reasoning task description steps of ChatGPT to construct a semantic map using a target recognition algorithm. They have designed a BERT-based parser module to parse the steps into initial BT. Then, the initial BT are dynamically expanded through a BT update algorithm to control the adaptive task execution of a robot. Lykov et al. [29] have proposed LLM-BRAIn, which is an AI-driven method based on an LLM. Existing research has demonstrated the potential of LLMs to generate behavioral models. However, their limited comprehension of complex logical relationships described in natural language often introduces ambiguity, inaccuracy, and instability into the resulting model. To overcome these shortcomings, this article introduces an LLM-driven framework for bidirectional FSM–BT conversion, designed to improve conversion efficiency, structural stability, and behavioral coherence.

\section{PROPOSED METHODOLOGY}
\subsection{Framework Overview}
\begin{figure}[!t] 
    \centering \includegraphics[width=\linewidth]{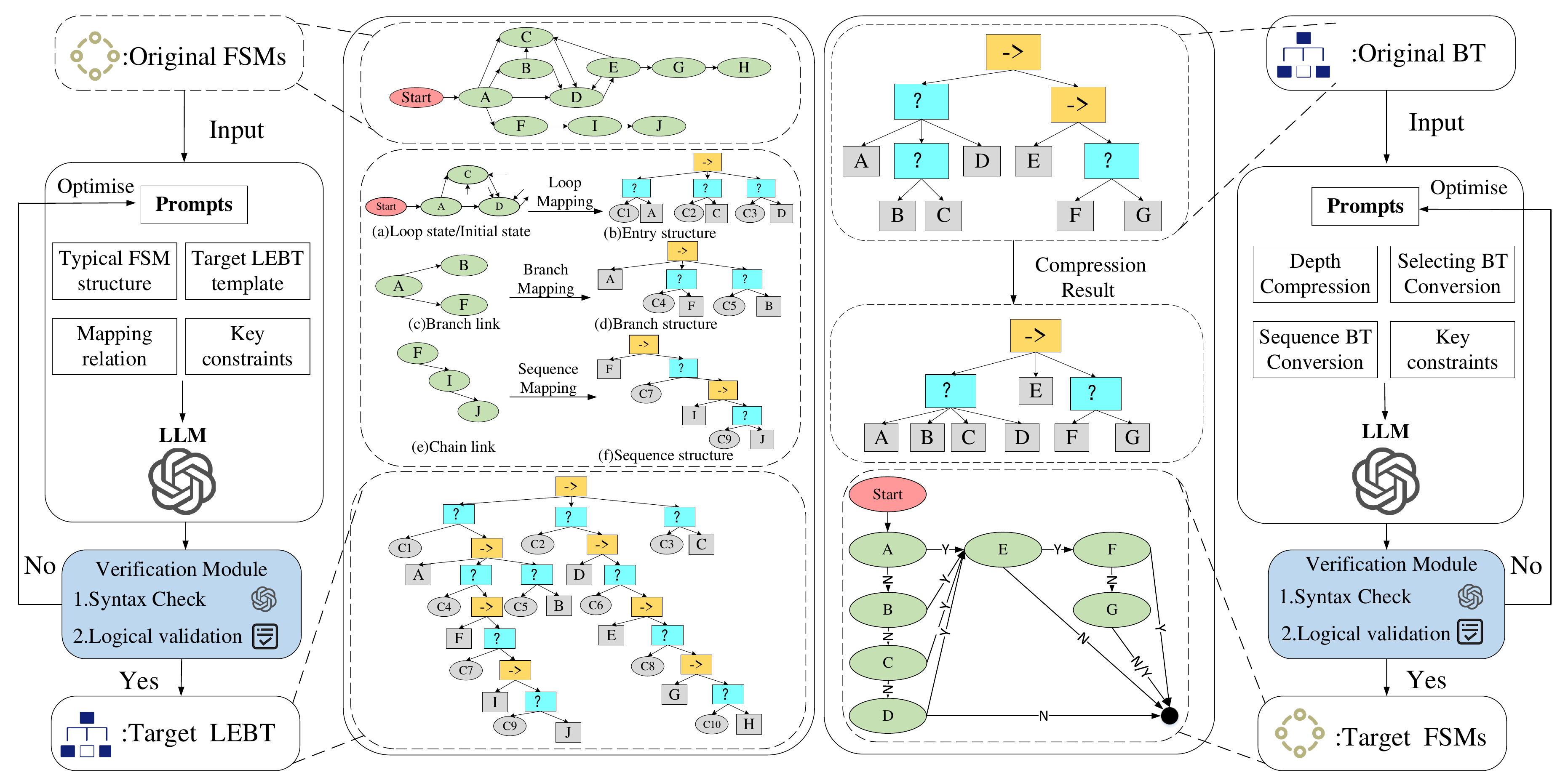} 
    \caption{LLM-based bidirectional conversion framework for FSM and BT behavioural models.} 
    \label{fig:fig1} 
\end{figure}
The proposed framework shown in Fig.\ref{fig:fig1} is initiated using a natural language conversion task to use LLM as a high-level behavioral model conversion engineer  specialized in bidirectional FSM–BT transformations. It comprises of three integrated components: the FSM-to-BT conversion component on the left, which performs model conversion through structural mapping; and the BT-to-FSM conversion component on the right, which performs model conversion through differentiated hierarchical conversion rules; at the bottom layer of the framework, a multi-level verification module is designed to ensure the structural correctness and behavioral consistency of the converted model. For models found to contain errors during validation, the LLM can be prompted to regenerate the model by optimizing the prompts (e.g., adding constraints or strengthening structural rules).

\subsection{FSM-to-BT Conversion Workflow}
This section describes the mapping template and workflow for the FSM-to-BT conversion.
\subsubsection{Typical Structure Recognition}
\begin{figure}[h] 
    \centering \includegraphics[width=\columnwidth]{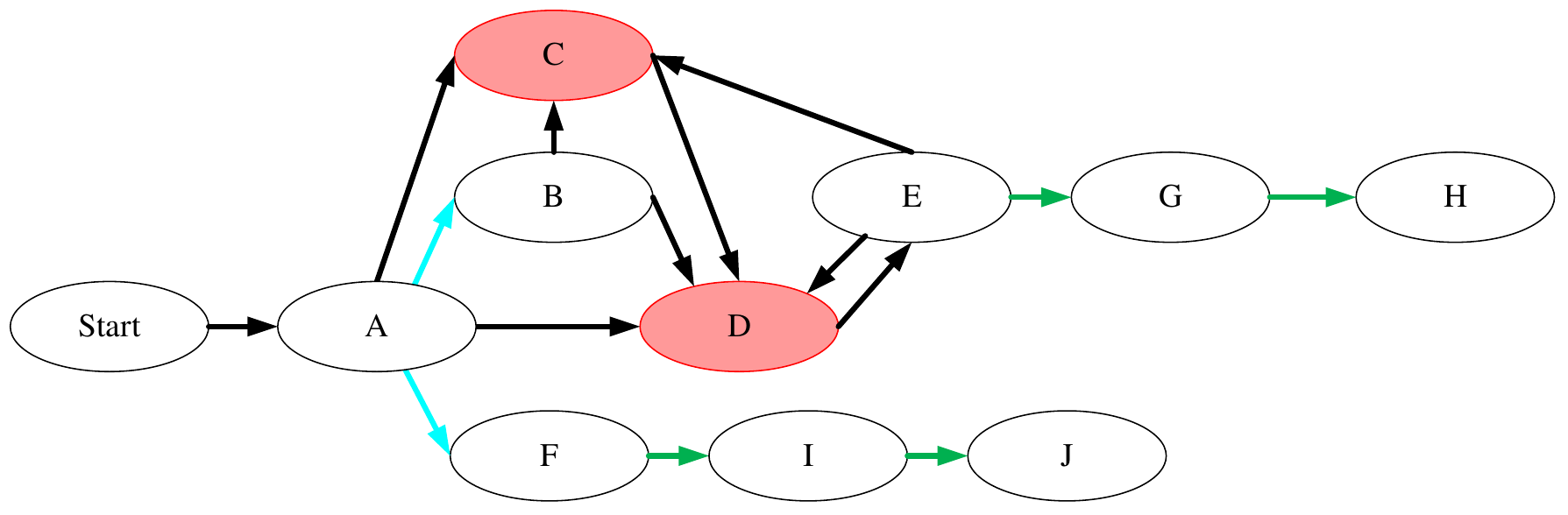} 
    \caption{Basic FSM structure.} 
    \label{fig:fig2} 
\end{figure}
There are four typical structures in FSMs: entry states, loop states, chain links, and branch links. The entry state is the state connected to the start state, such as state A in Fig. 2. Loop states are those with an in-degree greater than 1 (e.g., C and D in Fig.\ref{fig:fig2}). The state \textit{s} has $n$ ($0<n$) out-degrees. Should the $i$ ($1 < i \le n$) out-degrees connect to the ordinary state, the link formed between state $s$ and the ordinary states pointed to by these $i$ ($i=1$) out-degrees constitutes a branch link(A → B/F). Should \textit{i} out-degree connect to ordinary states, the link formed between state $s$ and the ordinary state pointed to by this single out-degree constitutes a chain link(E → G → H).

LEBT=($S{T_{entry}}$, $S{T_{branch}}$, $S{T_{seq}}$, $S{T_{sel}}$) is a quadruple.  $S{T_{entry}} = (r,ST_{sel}^{init},ST_{sel}^{loop})$ expresses the entry structure, where the root node $r$ is of sequence node type, with its child nodes comprising two sets of selector sub-trees: $ST_{sel}^{init}$ corresponding to the entry state set and $ST_{sel}^{loop}$ corresponding to the loop state set. This entry structure resides at the top level of the BT, serving to express the entry states and loop re-entry points of the FSM.
$S{T_{sel}} = (r,{c_l},{c_r})$ is the selector subtree, where the root node $r$ serves as the selector node. The left child node ${c_l}$ denotes the condition node, which maps the trigger events for state transitions. The right child node ${c_r} \in \{ {A_s},S{T_{branch}},S{T_{seq}}\} $ represents action nodes, branching structures, and sequential structures, designed to map states, branch links, and chained links within the FSM.
$S{T_{branch}}$=$(r,{A_s},ST_{sel}^1,ST_{sel}^2,...,ST_{sel}^k)$ represents the branching structure, where the root node $r$ is a sequence node. ${A_s}$ denotes the action node of state $s$, and each selector subtree in  $ST_{sel}^1,ST_{sel}^2,...,ST_{sel}^k$ corresponds to a link in the FSM where state $s$ behaves as an ordinary state, capturing the branching link within the FSM, where $k$ represents the number of links. When $k=1$ is a sequence structure $S{T_{seq}} = (r,{A_s},S{T_{sel}})$, it corresponds to chained links in the FSM. If a subsequent chain link exists, it is implemented by recursively nesting $S{T_{sel}}$ through the right subtree of $S{T_{seq}}$, producing ${c_r}(S{T_{sel}}) = S{T'_{seq}}$ .

\begin{figure}[h] 
    \centering \includegraphics[width=\columnwidth]{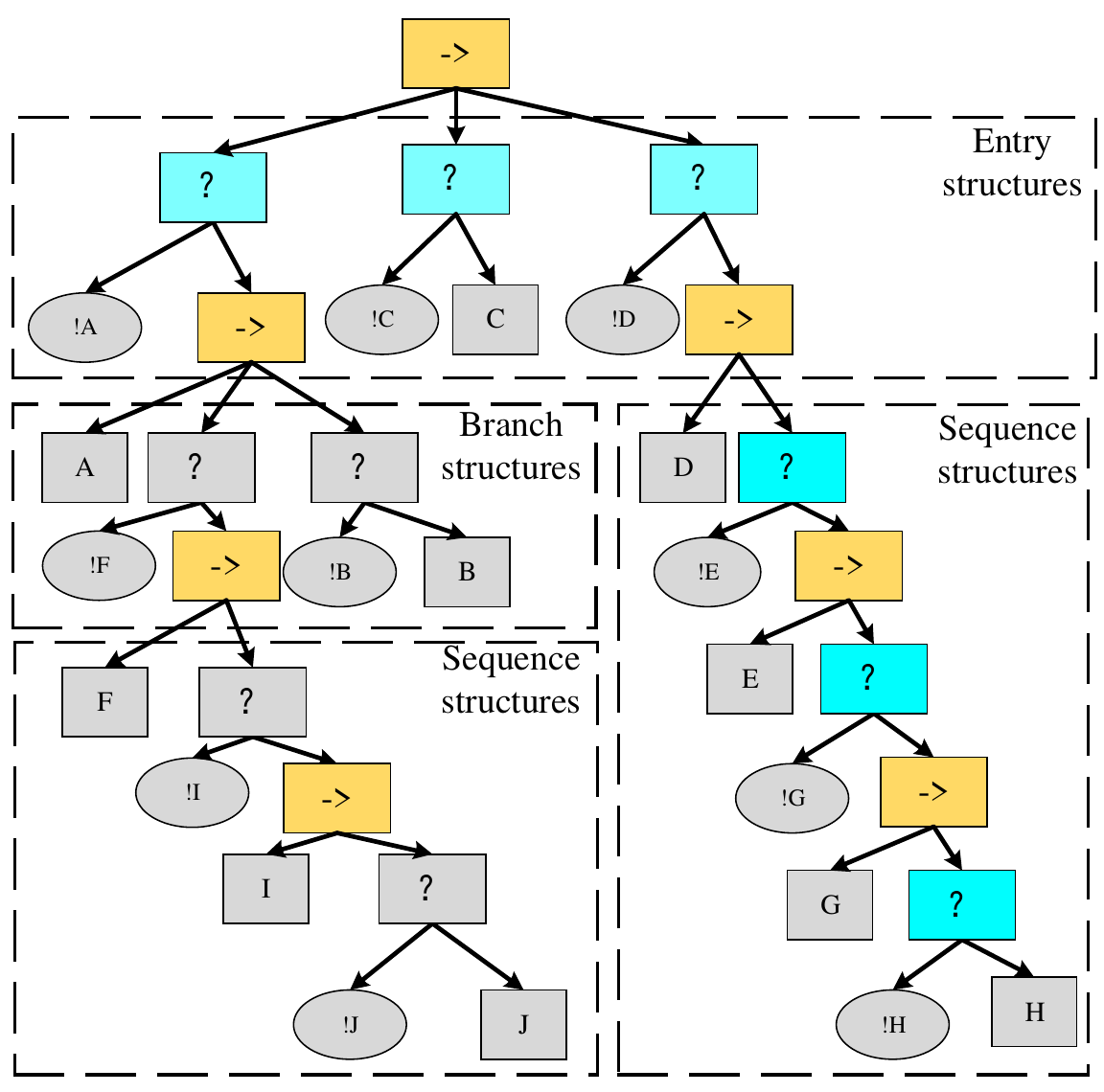} 
    \caption{LEBT structure.} 
    \label{fig:fig3} 
\end{figure}

\subsubsection{FSMs and LEBT Mapping Relationship}
Within the conversion framework, the states of the FSM are mapped to action nodes in the LEBT, whilst the events within the FSM are mapped to conditional nodes in the LEBT after being negated. Multiple events corresponding to the same action node are consolidated into a single conditional node. Furthermore, the entry states and loop states within the FSM are mapped to the selection subtrees within the LEBT entry structure. Branch links in the FSM are mapped to the branch structure in LEBT, whilst chained links are mapped to the sequence structure.

\subsubsection{LEBT Construction Process}First, the root node is directly connected to the entry structure, with the entry state and loop states sequentially added as selected subtrees, while duplicate state structures are removed. Second, each selector subtree's right child node is processed: If the corresponding state contains branch links, the node is converted into a branch structure. If chain links exist, they are converted into a sequential structure. Finally, the complete sequential structure is constructed recursively based on the order of the chained links.
\subsubsection{Prompts for FSM-to-BT Conversion}
The prompt module for FSM-to-BT conversion is structured into four units: Role, Task, Stepwise Conversion Procedure (SCP), and Constraints. This design systematically defines conversion tasks and clarifies constraints, thereby significantly enhancing the accuracy and stability of LLM outputs, as shown in Fig. 4. Within this framework, the LLM assumes the role of ‘Behavioral Model Expert’, whose main task is to achieve the correct conversion from FSM to BT, ensuring that the results generated exhibit consistent behavioral logic. The SCP first guides the model to recognize typical structures within the FSM; subsequently, it learns the characteristic structures of the proposed LEBT; then it completes the structural conversion based on the mapping relationship between the FSM and the LEBT structures; finally, it follows the LEBT construction process to complete the final BT construction. Key constraints include: prohibiting the introduction of any additional logic, strictly adhering to the defined conversion steps, and ensuring that the output is directly a complete BT description without any interpretations or extraneous text beyond the specified content\footnotemark[1]. 

\footnotetext[1]{https://github.com/zhoupeng-0425/FSM2BT}

\begin{figure}[h] 
    \centering \includegraphics[width=\columnwidth]{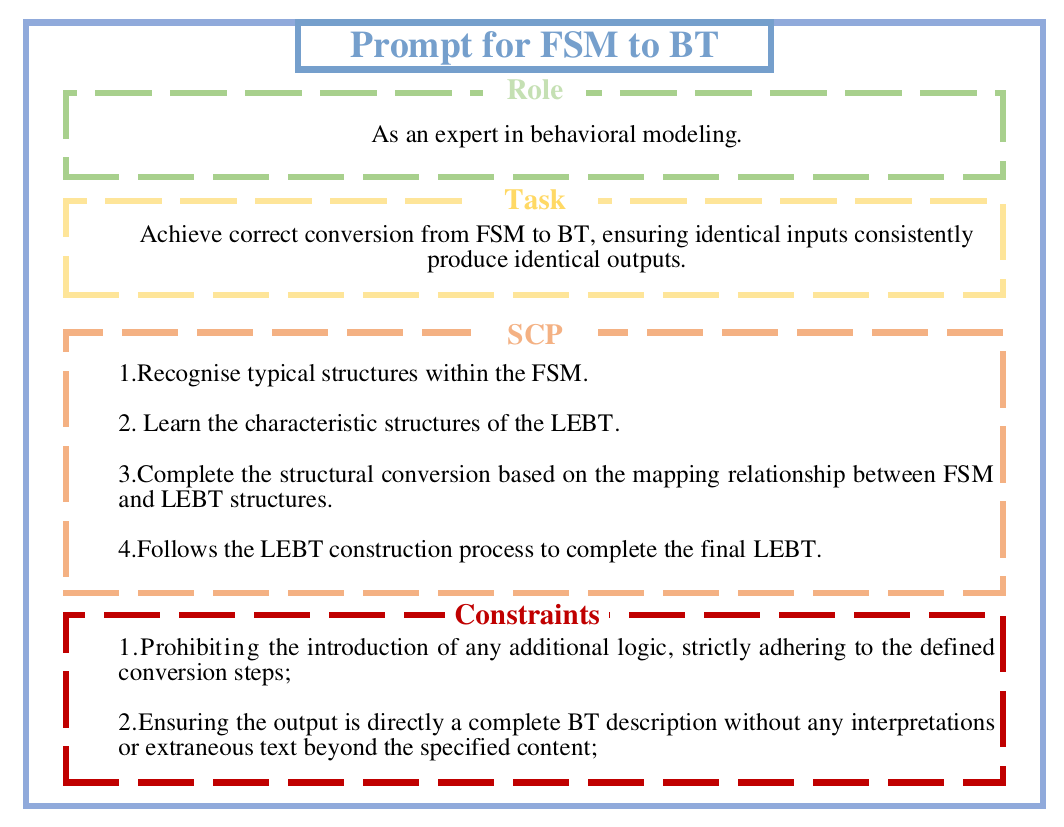} 
    \caption{Prompts for FSM-to-BT Conversion.} 
    \label{fig:fig19} 
\end{figure}

\subsubsection{FSM–LEBT Behavioral Equivalence Proof}
We prove behavioral equivalence by constructing a macro-step bisimulation [30] relationship between the pre- and post-transformation models, modeling both FSM and BT as labeled transition systems. If there exist $M = (S,\sum ,\delta ,{s_0},\alpha ,F)$ and $B = ({Q_B},\sum ,\Delta ,{q_0},\beta ,{F_B})$ such that there is a relation $R \subseteq S \times {Q_B}$ satisfying: 1. Initial preservation: $({s_0},{q_0}) \in R$; 2. Preservation of action : if $(s,q) \in R$, then $\alpha (s) = \beta (q)$; 3. Transition preservation: if $(s,q) \in R$ and $\delta (s,e) = s'$, then there exists $q' = \Delta (q,e)$ such that $(s',q') \in R$; if $(s,q) \in R$ and $\Delta(q,e) = q'$, then there exists $s' = \delta (s,e)$ such that $(s',q') \in R$ 4.Termination preservation: if $(s,q) \in R$, then $s \in F \Leftrightarrow q \in {F_B}$, then $M$ and $B$ are said to be macro-step bisimulation equivalents, denoted by $M \sim B$. Where, $M$ denotes an FSM, $S$ is a finite non-empty set of states; $\sum $ is a finite non-empty set of trigger events; $\delta :S \times \sum  \to S$ is the state transition function; $\delta $ is defined for every $(s,e) \in S \times \sum $ and produces a unique successor state for each input pair; $M$ is then referred to as a fully-defined deterministic FSM; ${s_0} \in S$ is the initial state; $\alpha :S \to Act$ is the state-action labeling function, denoting the observable actions executed when the system is in state $s$; $F \subseteq S$ is the set of terminal states. ${Q_B}$ is a finite set of configurations; a running configuration of an LEBT is defined as ${q_B} = (n,\rho )$, where $n \in N$ denotes the current position in the specification structure corresponding to the observable execution result; $\rho $ denotes the finite execution context required for a complete tick. In the LEBT specification discussed in this paper, as no parallel nodes, decorator nodes, or complex cross-tick local memory are introduced, all reachable running configurations form a finite set, denoted by ${Q_B}$; $\sum $ is the set of events; $\Delta :{Q_B} \times \sum  \to {Q_B}$ is the macro-step transition function; ${q_0} \in {Q_B}$ is the initial configuration; $\beta :{Q_B} \to Act$ is the observable action labeling function in the configuration; ${F_B} \subseteq {Q_B}$ is the set of terminal configurations. 

For any $s \in S$, define $\phi(s) \in Q_B$ as the stable configuration reached after the completion of a full LEBT tick. Specifically, $\phi(s)$ denotes the stable macro-step  configuration corresponding to state $s$, in which the action node $A_s$ is activated. The initial state of the FSM is $s_0$. Since the direct child of the LEBT root node is unique and corresponds to the entry structure node, and since the entry state $s_0$ is encoded as the initial entry branch in the entry structure, the first complete tick executed from the root node leads the system to the stable configuration corresponding to $s_0$. Denote this initial stable configuration by $q_0$. By the definition of $\phi$, we have $\phi(s_0)=q_0$, which establishes the initial-state condition.

Let $s \in S$ be arbitrary. By the LLMDUCF construction rule, the FSM state $s$ is uniquely mapped to the action node $A_s$, whose observable action label is exactly $\alpha(s)$. By the definition of $\phi$, $\phi(s)$ is the stable configuration in which $A_s$ is currently active. Since $\beta: Q_B \to Act$ assigns observable actions to LEBT configurations, the observable action of $\phi(s)$ is precisely the label of $A_s$. Hence,
\[
\beta(\phi(s)) = \alpha(s),
\]
which establishes the action-preservation condition.

Let $s \in S$ and $e \in \Sigma$ be arbitrary, and suppose that $s'=\delta(s,e)$ is defined. Since $M$ is a fully defined and deterministic FSM, the successor state $s'$ is uniquely determined. By the definition of $\phi$, $\phi(s)$ denotes the stable configuration in which the action node corresponding to the currently active state is $A_s$. Therefore, during the next complete tick starting from $\phi(s)$, only the successor structure associated with $A_s$ can be expanded. According to the LLMDUCF construction rule, for each state--event pair $(s,e)$, there exists a unique corresponding conversion unit. The success condition of this unit is precisely that the current state is $s$ and the input event is $e$, and its success result is the activation of the action node $A_{s'}$ corresponding to the target state $s'=\delta(s,e)$. For chained structures, the $ST_{seq}$ semantics ensures that, after a successful condition check, the execution proceeds to $A_{s'}$ in a fixed order. For branched structures, the $ST_{branch}$ semantics ensures that, among all candidate branches, exactly the branch corresponding to $(s,e)$ succeeds and activates $A_{s'}$. For loop cases, the $ST_{branch}$ structure ensures that the valid re-entry target is uniquely $A_{s'}$. Furthermore, since the mapping from state--event pairs to conversion units is unique, no two distinct conversion units can simultaneously correspond to the same state--event pair and lead to different target states. Therefore, a complete tick starting from $\phi(s)$ under the input event $e$ must, and can only, terminate in the stable configuration corresponding to $\phi (s')$. That is,
\[
\Delta (\phi (s),e) = \phi (s') = \phi (\delta (s,e)),
\]
which establishes the transition-preservation condition.

The set of final states $F$ of the FSM is mapped to the set of final configurations $F_B$ under the LEBT macro-step semantics. Specifically, if $s \in F$, then the corresponding action node $A_s$ either does not derive any further valid successor behavior in the construction, or its corresponding stable configuration is explicitly designated as a terminating configuration. Conversely, if $s \notin F$, then there exists a valid successor structure triggered by some event in the construction, and hence the stable configuration corresponding to $s$ is not a terminating configuration. Therefore, we have
\[
s \in F \Leftrightarrow \phi(s) \in F_B,
\]
which establishes the final-state preservation condition.

In summary, the mapping $\phi$ preserves the initial state, observable actions, transition behavior, and final states between the FSM and the LEBT macro-step semantics. Hence, the FSM and the LEBT are bisimilar at the macro-step level. Consequently, the LEBT constructed by LLMDUCF is behaviorally equivalent to the original FSM. The FSMs supported by the framework are deterministic finite transition systems with a finite set of states, where each state--input pair leads to a unique successor state.

\subsection{BT-to-FSM Conversion Workflow}
This section presents the workflow for the BT-to-FSM conversion.

\subsubsection{BT Depth Compression strategy}If the main tree and sub-trees of a BT share the same type of control node, the sub-tree control nodes can be removed, and the sub-tree structure can be promoted to the main tree, as shown in Figs. 5 and 6. The deep compression strategy not only eliminate redundant BT control structures to simplify tree structures but also transform BT into alternating sequential and selection node control structures to accommodate differentiated hierarchical conversion rules.
\begin{figure}[h] 
    \centering \includegraphics[width=\columnwidth]{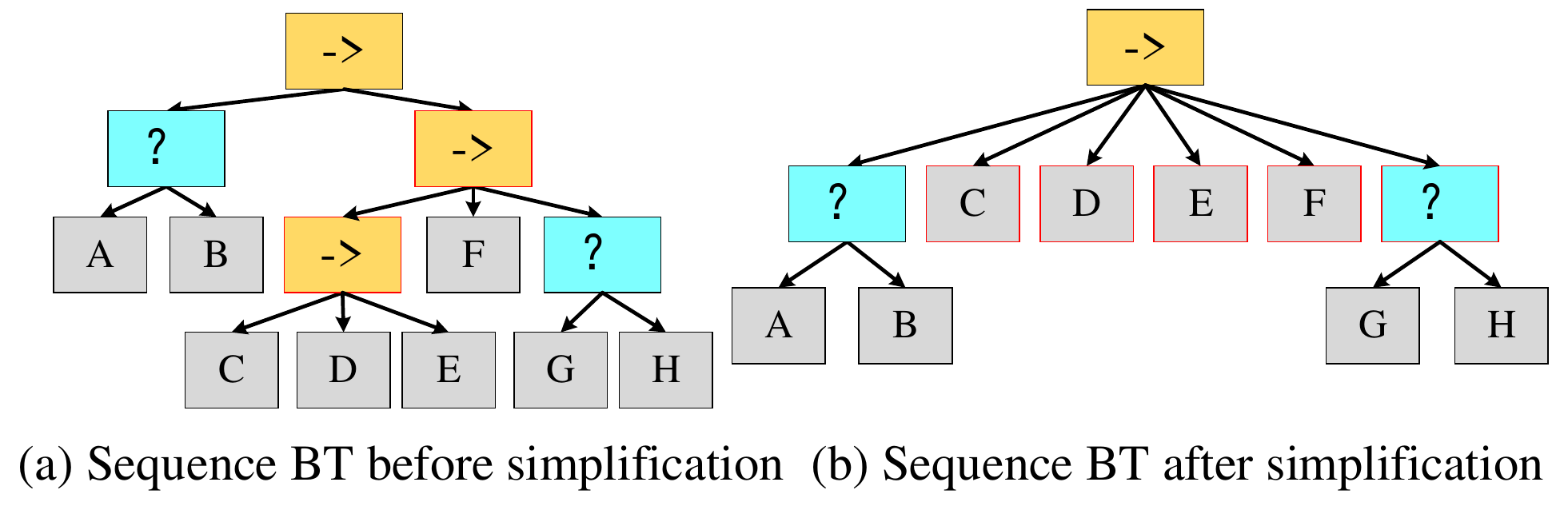} 
    \caption{Simplified flowchart for sequence BT.} 
    \label{fig:fig4} 
\end{figure}
\begin{figure}[h] 
    \centering \includegraphics[width=\columnwidth]{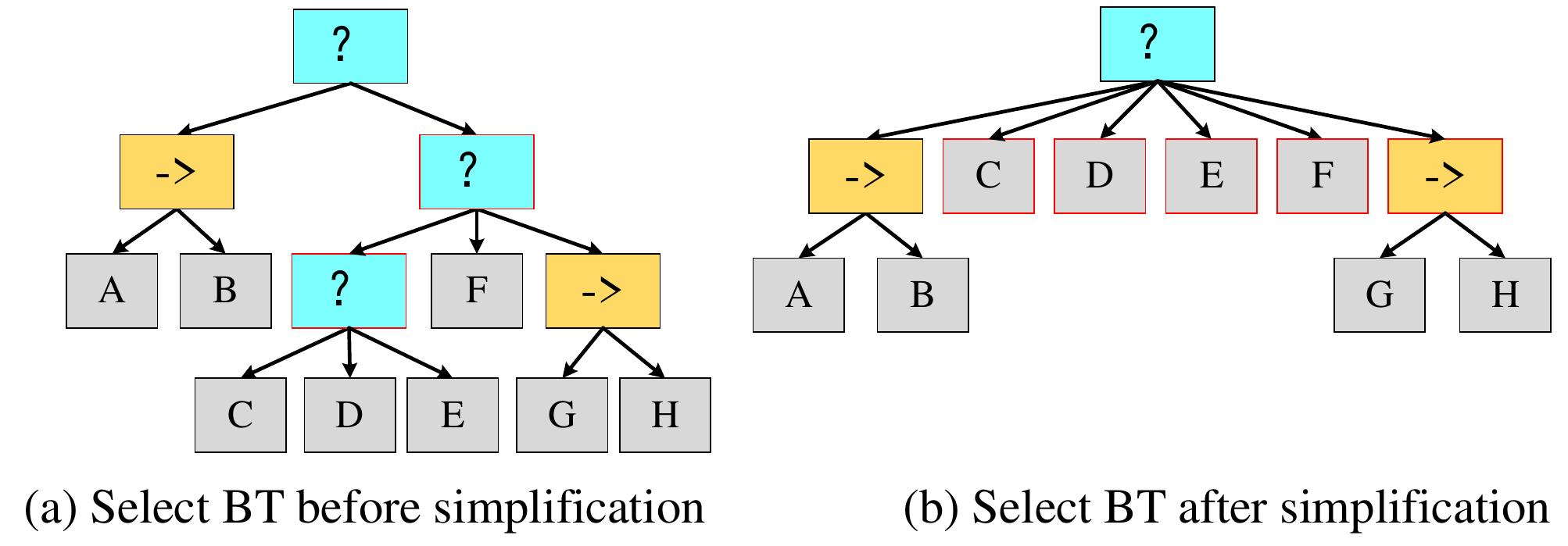} 
    \caption{Simplified flowchart for selector BT.} 
    \label{fig:fig5} 
\end{figure}

\subsubsection{Differentiated Hierarchical Conversion Rules}The differentiated hierarchical conversion rules first divide the BT into conversion units every three layers. Following deep compression processing, each conversion unit retains structural consistency. Subsequently, every conversion unit is transformed into a sub‑FSM according to the respective execution logic of sequence nodes and selector nodes. Finally, all generated sub-FSM are integrated into a complete FSM. The conversion results for sequence BT and selector BT are shown in Figs 7 and 8.

\begin{figure}[h] 
    \centering \includegraphics[width=\columnwidth]{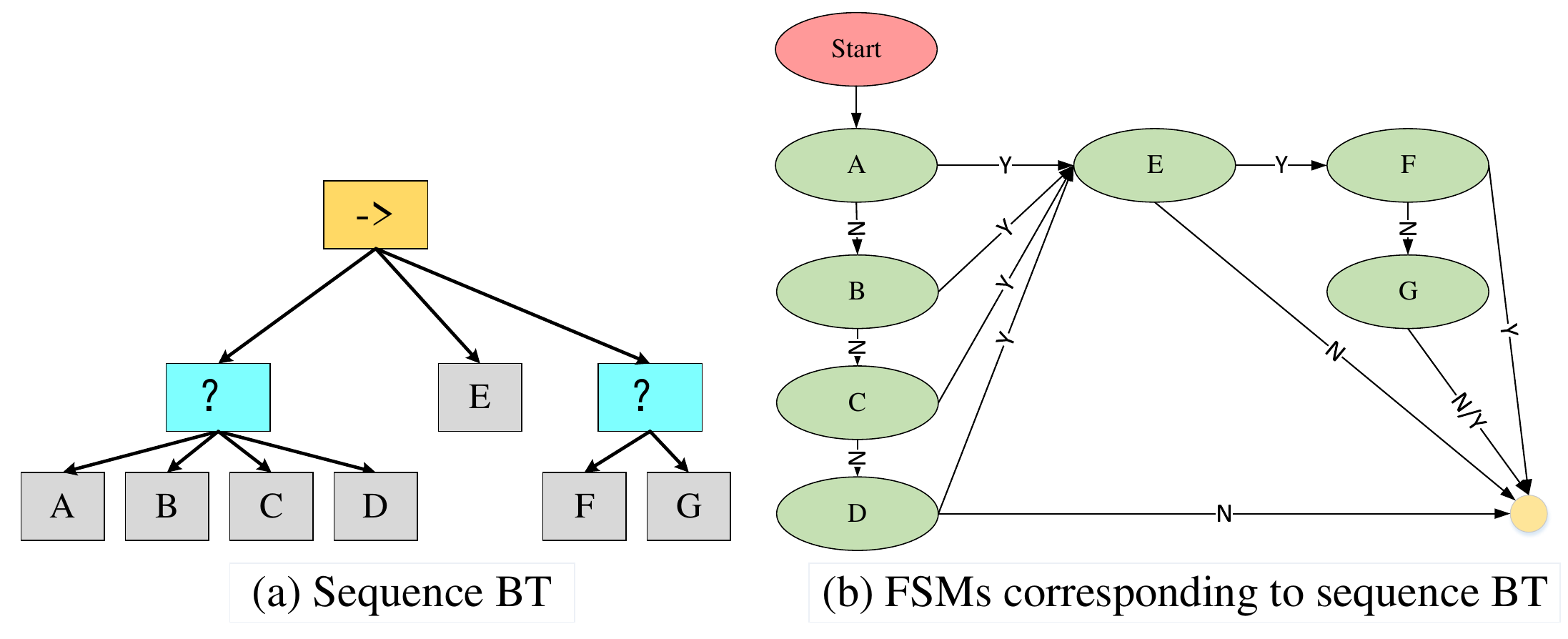} 
    \caption{Structure Diagram of Sequence BT Conversion to FSM.} 
    \label{fig:fig6} 
\end{figure}

\begin{figure}[h] 
    \centering \includegraphics[width=\columnwidth]{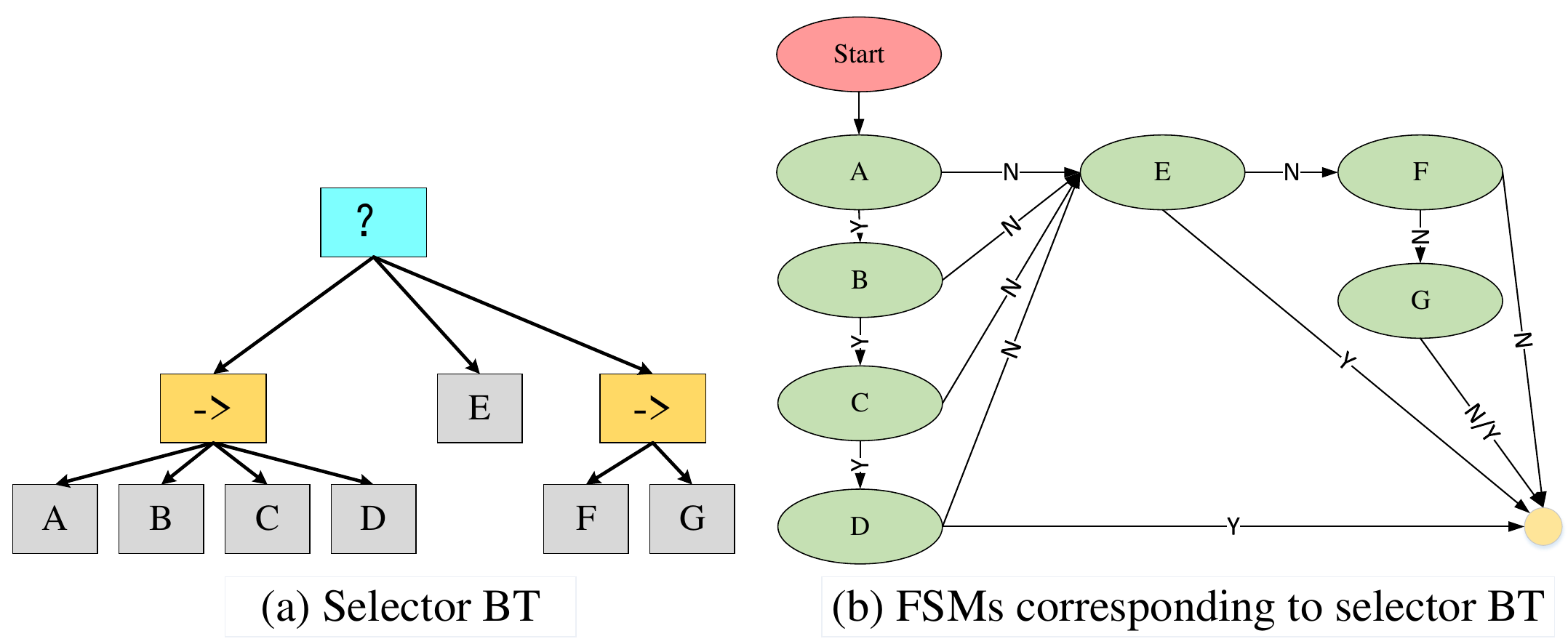} 
    \caption{Structure Diagram of Selector BT Conversion to FSM.} 
    \label{fig:fig7} 
\end{figure}

\subsubsection{FSM Construction Process}First, map the leaf nodes of the BT to states, and map the transition units to child FSM; Second, convert all transition units into their corresponding child FSM according to the established transition rules; Finally, integrate the child FSM generated from all transition units to form a complete FSM. 

\begin{figure}[h] 
    \centering \includegraphics[width=\columnwidth]{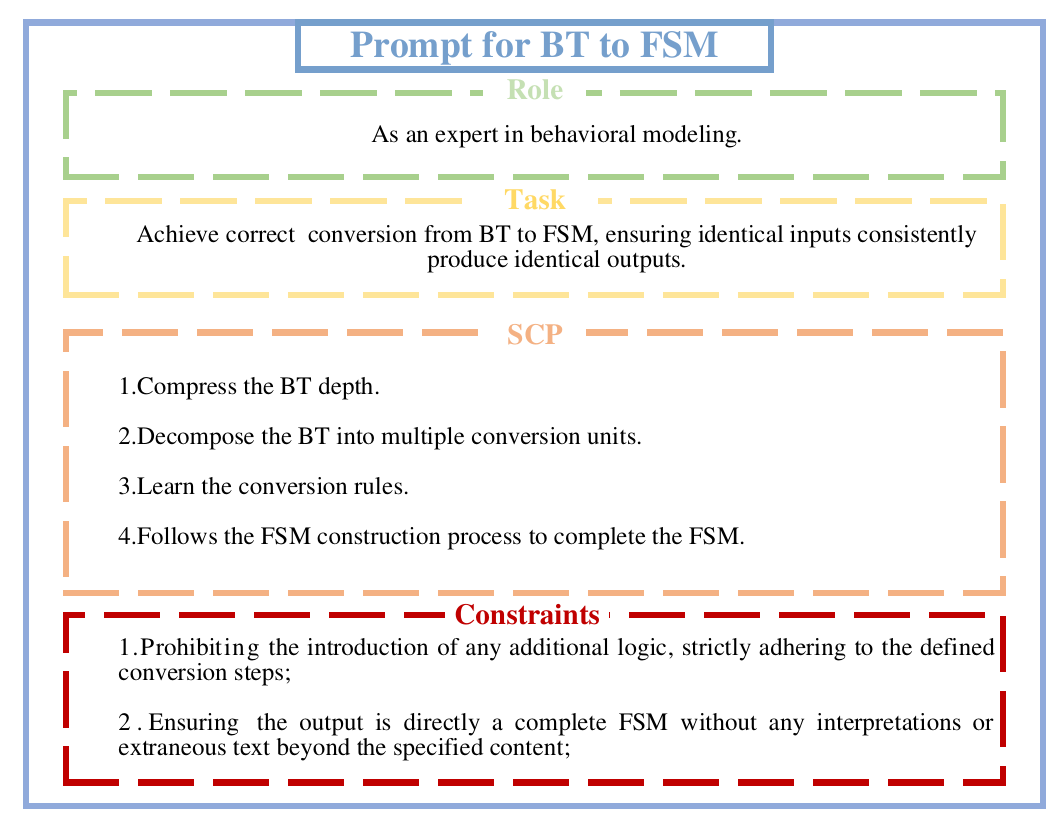} 
    \caption{Prompts for BT-to-FSM Conversion.} 
    \label{fig:fig20} 
\end{figure}

\subsubsection{Prompts for BT-to-FSM Conversion}The prompt for converting BT to FSM is illustrated in Fig. 9. Here, the LLM assumes the role of a ‘Behavioral Model Expert’, with its main task being to achieve the correct conversion from BT to FSM, ensuring that the results generated exhibit consistent
behavioral logic. The SCP first guides the model to compress the depth of BT and decompose the BT into multiple conversion units; second, it learns the conversion rules proposed in this paper; finally, it completes the FSM construction by adhering to the FSM construction process. Crucial constraints include prohibiting the introduction of any additional logic; strictly adhering to the defined conversion steps; and ensuring that the output is a complete FSM description without any explanatory text or extraneous content beyond the specified requirements.

\subsubsection{BT-FSM Behavioral Equivalence Proof}
This paper implements a direct conversion from BT to FSM based on execution semantics, and demonstrates from a linguistic-semantic perspective that the execution semantics of this FSM are trajectory-equivalent to those of the original BT. For nested sequence structures of type ${Seq(a,Seq(b),c) \equiv Seq(a,b,c)}$, both visit child nodes in exactly the same left-to-right order and halt upon the first unsuccessful execution; consequently, the leaf node visit sequences, atomic observation sequences, return values, and memory updates are consistent. If execution is in progress, the control remains at the same leaf node. Similarly, the same holds for ${Sel(a,Sel(b),c) \equiv Sel(a,b,c)}$; therefore, the semantic integrity of the depth-first compression method is preserved. The sequential conversion unit visits child nodes in sequence; if a child node is a leaf node and its execution succeeds, it transitions to the next node, continuing until the last child node executes successfully, at which point the entire conversion unit is deemed to have executed successfully. If a leaf node fails, the conversion unit is deemed to have failed; if execution is in progress, it resumes from the current node on the next iteration. If the child node is a selection sub-tree, the child nodes are visited sequentially. If a child node is a leaf node or the execution of a child transformation unit fails, the process moves to the next node in the selection sub-tree, until the last child node fails, at which point the entire transformation unit is deemed to have failed. If a leaf node in the selection sub-tree executes successfully, the process moves to the next leaf node in the transformation unit; this is consistent with the logic of the generated FSM. By the same way, it follows that the selector conversion unit is equivalent to the FSM generated by the transition; therefore, BT is equivalent to FSM. The BT supported by the framework are normalized behavior trees that contain only selector nodes, sequence nodes, and leaf nodes.

\subsection{Verification Module}
LLMDUCF employs a two-stage verification mechanism to ensure the reliability of the generated models. First, an LLM-based verifier checks the syntax and structure of the generated BT/FSM against predefined grammar rules and structural constraints, identifying errors such as dangling states or nodes, duplicated structures, missing nodes, invalid branches, and incomplete coverage. Second, NuSMV\footnotemark[2] is used to verify the behavioral consistency between the source and generated models. The original FSM/BT and the converted BT/FSM are abstracted into NuSMV-compatible finite-state transition systems, and formal properties are constructed to check whether the converted model preserves the execution behavior of the source model. When inconsistencies are detected, the feedback is returned to the LLM to guide regeneration through prompt refinement and additional structural constraints, thereby improving the correctness and stability of bidirectional FSM-BT conversion.

\section{EXPERIMENTS AND VALIDATION}
To quantitatively evaluate the performance and generalizability of our proposed framework, we conducted experiments in three distinct hypothetical scenarios, all scenarios using the same text input. This multi-scenario evaluation includes a basic symbolic scenario for foundational verification, a robotic manipulation scenario for physical interaction tasks, and a game AI scenario for dynamic decision-making . The performance of our framework is systematically compared against a normal LLM (NLLM) without SCP and traditional translation approaches.

\subsection{Experimental Settings}
\subsubsection{Experimental Dataset}We constructed a behavioral-model dataset spanning multiple application scenarios. The dataset combines models randomly generated under formally defined syntactic constraints with behavioral models collected from open-source projects. Covers diverse node-naming conventions, nesting depths, branching and looping patterns, model sizes, and domain-specific semantic descriptions, enabling a representative and rigorous evaluation of the proposed framework across different behavioral-model structures and application contexts. The dataset contains only the source models to be converted; the generated target models are not included in the dataset and are evaluated separately using NuSMV to verify their behavioral consistency with the corresponding source models.

Scenario 1: The basic symbol scenario verifies the fundamental correctness and completeness of the formal core structures (such as sequence, selector, and loop) during the mutual conversion between FSM and BT.
Scenario 2: The robotic grasping scenario tests the framework's performance under typical physical interactions and embodied decision-making tasks. Scenario 3: The game AI scenario evaluates the framework's generalizability and practicality within autonomous systems requiring high dynamics, real-time responses, and non-deterministic decision-making. The experimental scenarios described above are closely related to AISs. Robot grasping scenarios represent object selection, navigation, grasping, and placement tasks performed by service robots, home robots, and smart logistics devices in human-centered environments. Game AI scenarios represent interactive entertainment systems, in which non-player characters must possess the ability to make dynamic decisions, track targets, patrol, evade, and adaptively switch behaviors. These two scenarios cover embodied consumer electronics devices and software-based consumer entertainment systems, respectively, and reflect typical modeling and transformation requirements for autonomous behavior in the consumer electronics field. 

Each scenario includes 200 BT and 200 FSM. The dataset\footnotemark[1] contains approximately 350 behavioral models from open-source projects. For example, the behavioral model shown in Fig. 14 is a model from the open-source robotics library\footnotemark[3]. The model implements the core task logic for autonomous cargo selection, path planning, target cargo grasping, and cargo placement at fixed locations in industrial logistics handling scenarios. The research covers boundary conditions, naming conventions, and structural constraints commonly found in actual robotic systems. For game AI scenarios, as shown in Figure 20, the NPC control code libraries are derived from open-source game engines\footnotemark[4]. These models implement the dynamic decision-making logic of predator-prey game agents, including patrol and exploration, target detection and tracking, prey capture, foraging, and danger avoidance/escape. The behavioral models for the remainder of the dataset were generated randomly. BT are composed of basic node types, with depth and breadth constrained to a range of 2 to 10. FSM incorporate chain, branch, and loop structures, with a number of states ranging from 2 to 30. According to expert studies, BT structures deeper than 8 layers lead to a sharp increase in comprehension difficulty, while FSM with more than 25 states become significantly harder to understand, test, and extend.

\footnotetext[2]{https://nusmv.fbk.eu}
\footnotetext[3]{https://github.com/ethz-asl/bt$\_$fsm$\_$comparison}
\footnotetext[4]{https://github.com/NUDTQI/PartoPrey-BT-RL}

\subsubsection{Experimental Environment} 
Scenario 1 employs manual inspection to verify the accuracy of the conversion. In Scenario 2, some models were simulated using the Robot Operating System (Humble ROS 2) and Gazebo 11 platform. In Scenario 3, some behavioral models
were simulated using the open-source
Predator-Prey Game platform. NuSMV (2.6.0-win64) is a symbolic model checker designed for formal verification of finite-state systems. It is used here to verify the converted FSM and BT models, ensuring that they satisfy behavioral consistency properties such as reachability, conversion completeness, deadlock freedom, and termination consistency. The experimental hardware configuration
included an NVIDIA RTX 3060 graphics card and 16GB
of memory.

ROS is an open-source robotics software framework that provides hardware abstraction, device drivers, messaging, etc., and supports modular development. Gazebo is a stand-alone 3D robotics simulation environment that focuses on high-fidelity physics and sensor simulations. The robot grasping task is implemented using the Gazebo simulator, as shown in Fig. 10.

\begin{figure}[h] 
    \centering \includegraphics[width=\columnwidth]{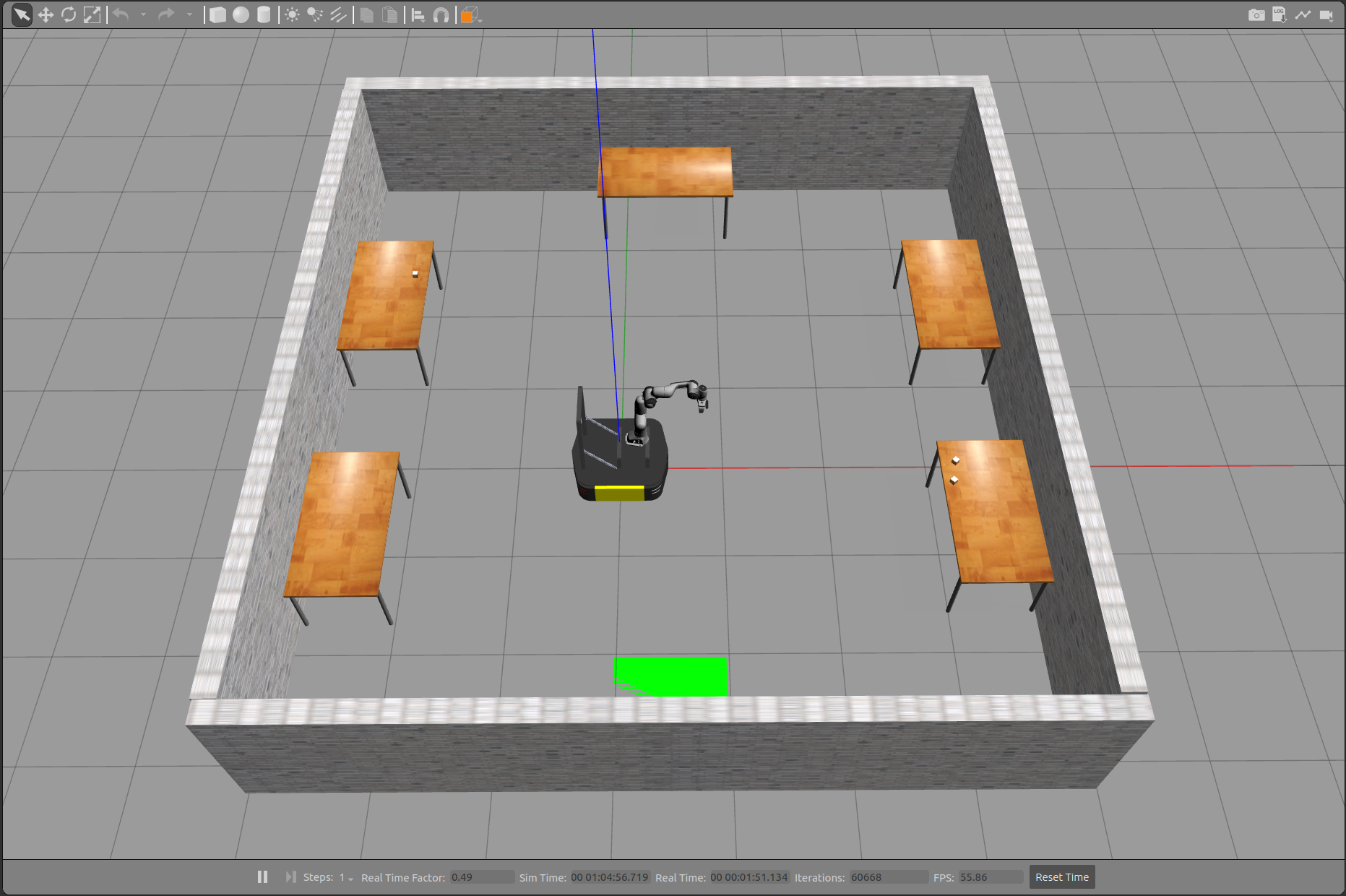} 
    \caption{Gazebo simulator.} 
    \label{fig:fig8} 
\end{figure}

A predator–prey game simulation platform based on C++ is used to simulate the predator-prey process, in which the behaviors of the predator and the prey are controlled by AI. The platform provides an AI modeling interface, FSM and BT behavioral modeling, and a large number of basic actions such as roaming, tracking, capturing, and maneuvering. The pursuer capture task is implemented on this platform, as shown in Fig. 11. 

\begin{figure}[h] 
    \centering \includegraphics[width=\columnwidth]{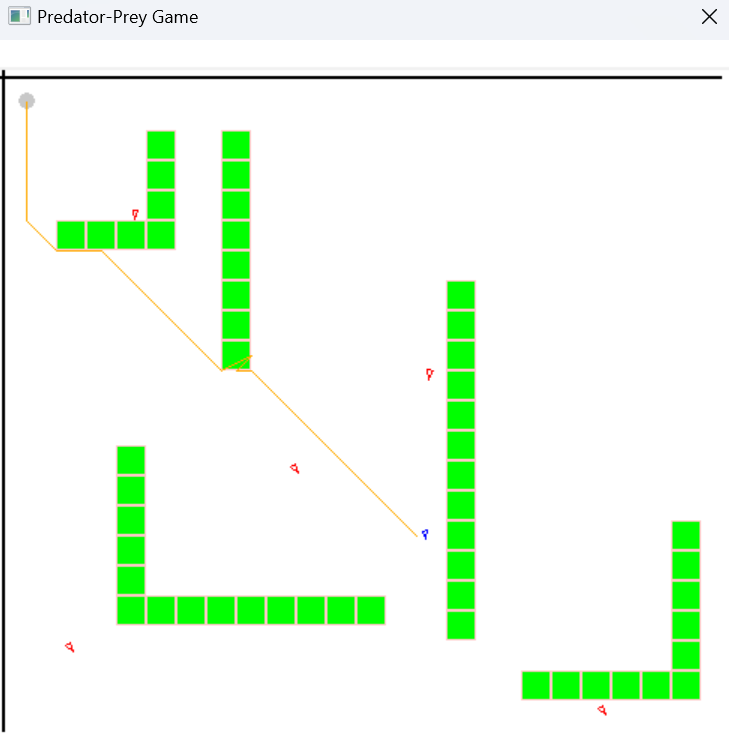} 
    \caption{Predator-Prey Game Simulation Platform.} 
    \label{fig:fig8} 
\end{figure}

\subsubsection{Experimental Setup for LLM}
In this study, we conducted experiments using three different LLMs: ChatGPT (gpt-4o), DeepSeek (deepseek-reasoner) and Qwen (qwen3-max-thinking). The reason  for this selection is as follows: 1. All three are reasoning-capable large models with native support for Chain-of-Thought logical reasoning, making them well-suited to the structured transformation requirements of behavioral models; 2. They encompass both internationally dominant closed-source models and domestically developed open-source models, validating the framework's cross-model universality; 3. All provide stable API interfaces, supporting reproducible experiments with fixed hyperparameter, and represent the main choices for current behavioral modeling research in the field. All LLMs were accessed via the OpenAI API (version 1.30.0+), with the following key parameters fixed to ensure reproducibility: the temperature was set to 0.1 and the random seed was fixed; the Top-p to 0.9 to enhance output stability; the max retries was set to 3, and the request timeout was set to 180 seconds. The error output verified by the validation module may be reprocessed with optimized prompts.

\subsection{Evaluation Metrics}
Evaluating the behavioral model transformation framework from five perspectives: Conversion Accuracy, Path Coverage, Modularity, Readability, and Input Representation Generalization Ability.

\subsubsection{Conversion Accuracy}Conversion accuracy is used to evaluate the reliability of the framework in performing behavioral model conversion tasks repeatedly. The conversion accuracy is defined as:

\[
Accuracy = \frac{N_{first-pass-success}}{N_{model}}
\]

where ${N_{model}}$ represents the total number of test cases; and ${N_{first-pass-success}}$  denotes the number of test cases for which the initially generated model passes both structural validation and NuSMV-based formal verification, without any retry or regeneration.

\subsubsection{Behavioral Consistency}Behavioral consistency evaluates whether the converted model preserves the execution semantics of the source model. In this work, formal verification is used to check whether the converted model satisfies the following properties: all critical states, action nodes, and behavioral phases in the source model are reachable in the converted model; each valid transition relation or behavioral path in the source model has a corresponding execution path in the converted model; no additional critical behavior sequence absent from the source model is introduced; no deadlock exists in any non-terminal state; and, under the same input events or condition sequences, the source and converted models reach consistent success, failure, or terminal states. A converted model is regarded as behaviorally consistent if all these properties are satisfied.

\subsubsection{Modularity}This paper compares the addition, deletion, and modification operations of behaviors within decision structures, focusing on their computational complexity and edit distance. Computational complexity refers to the computational resources and the number of steps required to perform addition, deletion, or modification operations on nodes or states within a BT or FSM. The edit distance denotes the minimum number of edit operations required to transform one structure into another target structure.

\subsubsection{Readability}Comparing the scalability and maintainability of the behavior model from a readability perspective, treat behavior actions (represented as nodes in BT and states in FSM) as variables to calculate the number of graphical elements within the behavior model structure.Graphical elements constitute all visual components within the behavioral model, including nodes (or states) and edges (or events) that can be intuitively represented. Active elements denote core components within the behavioral model that are directly manipulable (add, delete, modify), serving as the fundamental elements that influence the behavioral model.

\subsubsection{Input Representation generalizability } The ability of a method to accurately understand the semantic meaning of a behavior and correctly perform model transformation when the same behavioral model is presented in different input formats or representations. This ability is reflected mainly in the method’s adaptability to different input definitions, input formats, and transformation conventions.

\subsection{Baselines}The experimental setup for NLLM is identical to that of LLMDUCF, employing prompts without SCP. Traditional behavior model conversion methods, including bidirectional structural mapping (BSD) [15] and CHDS–BT [16] method, are also selected to compare with LLMDUCF.

\subsection{Performance of BT-to-FSM Conversion}
\begin{figure}[h] 
    \centering \includegraphics[width=\columnwidth]{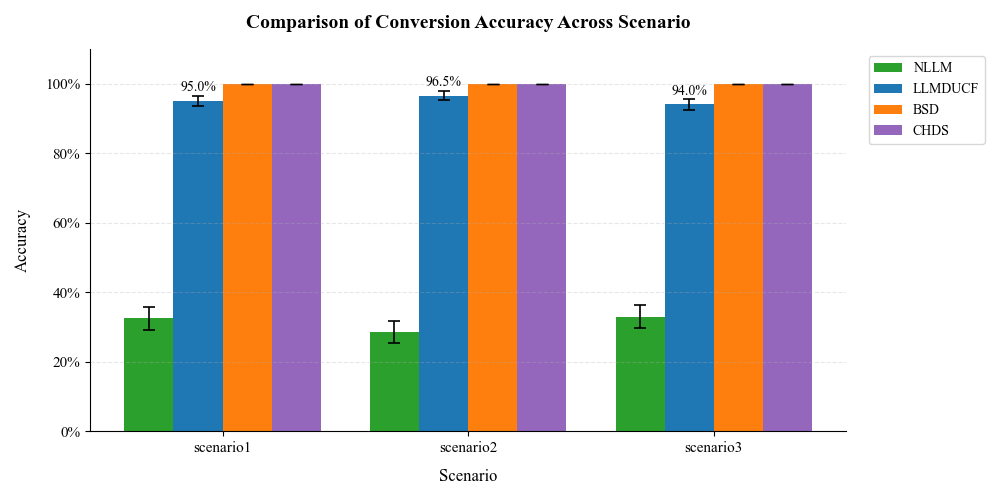} 
    \caption{Comparison of Conversion Accuracy Across Scenario.} 
    \label{fig:fig8} 
\end{figure}

In all evaluation scenarios, traditional conversion methods BSD and CHDS-BT achieved conversion accuracy of 100$\%$, due to their reliance on fixed conversion rules, as shown in Fig. 12. The proposed LLMDUCF, when implemented with ChatGPT, significantly outperformed the NLLM in all three scenarios (Scenario 1: $95.70\%{ _{\pm4.21}}$, Scenario 2: $96.95\%{ _{\pm3.61}}$, Scenario 3: $93.65\%{ _{\pm4.63}}$), where the confidence interval is 95$\%$. This performance improvement stems primarily from two key design elements: first, the depth of the behavior tree is reduced through deep compression strategy, forming an alternating control structure; second, the proposed differential hierarchical transformation rules partition complex behavior trees into multiple transformation units for separate processing. This approach enhances conversion accuracy while effectively reducing the number of generated sub-FSM.

\begin{figure}[h] 
    \centering \includegraphics[width=\columnwidth]{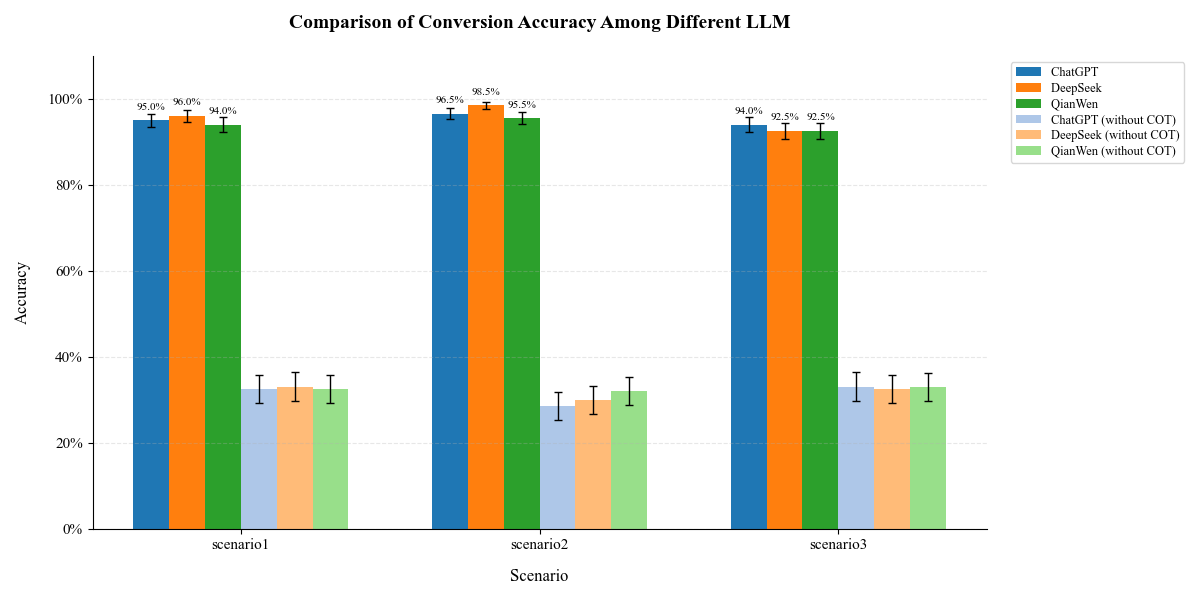} 
    \caption{Comparison of Conversion Accuracy Among Different LLM.} 
    \label{fig:fig8} 
\end{figure}

\begin{table}[h]
    \centering
    \caption{Conversion Accuracy with 95$\%$ confidence intervals}
    \label{tab:fsm_conversion}
    \begin{tabular}{l c c c}
        \toprule
        \textbf{} & \textbf{scenario1} & \textbf{scenario2} & \textbf{scenario3} \\
        \midrule
        ChatGPT  & $95.70{ _{\pm4.21}}$ &$96.95{ _{\pm3.61}}$ & $93.65{ _{\pm4.63}}$  \\
        DeepSeek &$96.35{ _{\pm3.00}}$ &$97.80{ _{\pm2.36}}$ & $90.65{ _{\pm5.58}}$ \\
        Qwen  &$94.10{ _{\pm4.26}}$ &$94.75{ _{\pm4.99}}$ & $93.55{ _{\pm4.59}}$ \\
        ChatGPT (without SCP) &$32.10{ _{\pm9.27}}$ &$28.70{ _{\pm8.16}}$ & $31.90{ _{\pm9.38}}$  \\
        DeepSeek (without SCP) &$32.80{ _{\pm9.10}}$ &$31.50{ _{\pm8.14 }}$ & $31.00{ _{\pm8.00}}$  \\
        Qwen (without SCP) &$32.60{ _{\pm9.34}}$ &$32.45{ _{\pm8.14}}$ & $31.05{ _{\pm8.24}}$  \\
        \bottomrule
    \end{tabular}
\end{table}

In simulation experiments, we evaluated the performance of different LLMs in converting BT to FSM, with the results shown in Fig. 13. In all test scenarios, LLMs employing SCP prompts achieved significantly higher conversion accuracy than baseline models without this strategy, maintaining accuracy consistently between 90.65$\%$ and 97.80$\%$. This outcome demonstrates that SCP prompts, specifically designed for model conversion tasks, effectively guide LLMs in comprehending the semantic structure of BT, thereby enhancing conversion precision. Specifically, the main LLM that employs SCP exhibited robust performance in all scenarios, consistently achieving accuracy rates above 90$\%$ with minimal fluctuation. This confirms the strong versatility and compatibility of the model with the prompt framework. Moreover, the SCP strategy achieved high-precision, low-variance conversion results in three distinct test scenarios, further demonstrating its cross-scenario robustness in performance enhancement. In summary, the LLMDUCF framework achieves high-precision and stable conversion outcomes in various LLM and task scenarios by integrating SCP prompts, demonstrating excellent adaptability to mainstream models. SCP prompts successfully bridge the semantic gap between general-purpose LLM and structured formal model conversion tasks.

\begin{table}[h]
    \centering
    \caption{Ablation Experiments}
    \label{tab:fsm_conversion}
    \begin{tabular}{l c c c c}
        \toprule
        \textbf{} & \textbf{scenario1} & \textbf{scenario2} & \textbf{scenario3} &  \\
        \midrule
        Base &$32.10{ _{\pm9.27}}$ &$28.70{ _{\pm8.16}}$ & $31.90{ _{\pm9.38}}$  \\
        Base+M1 &$37.05{ _{\pm9.02}}$ &$32.55{ _{\pm8.99}}$ & $33.85{ _{\pm9.14}}$ \\
        Base+M2 &$83.40{ _{\pm7.08}}$ &$86.85{ _{\pm6.89}}$ & $84.50{ _{\pm7.35}}$  \\
        Base+M1+M2 &$95.70{ _{\pm4.21}}$ &$96.95{ _{\pm3.61}}$ & $93.65{ _{\pm4.63}}$ \\
        \bottomrule
    \end{tabular}
\end{table}

To elucidate the independent and synergistic effects of each module within the SCP framework of the LLMDUCF method, we conducted systematic ablation experiments. The conversion accuracy results under each module configuration are summarized in Table II. M1 comprises the deep compression strategy, while M2 involves differentiated hierarchical transformation rules. The experiments demonstrate that enabling the deep compression algorithm alone yields limited performance gains. This module primarily assists large language models in achieving correct conversions by simplifying behavior tree structures with depths of 3 to 4 layers; however, its capability to handle more complex nested structures remains constrained. Enabling differentiated hierarchical conversion rules alone yielded more pronounced performance gains. Its core mechanism involves decomposing the complex BT into multiple simple conversion units for separate processing. However, this rule is primarily suited to tree structures alternately controlled by sequential and selector nodes. The differentiated hierarchical conversion rules constitute the core module for enhancing overall conversion performance. The depth compression algorithm not only generates independent performance gains, but also synergizes with these conversion rules to further optimize the final conversion results.

\begin{figure}[h] 
    \centering \includegraphics[width=\columnwidth]{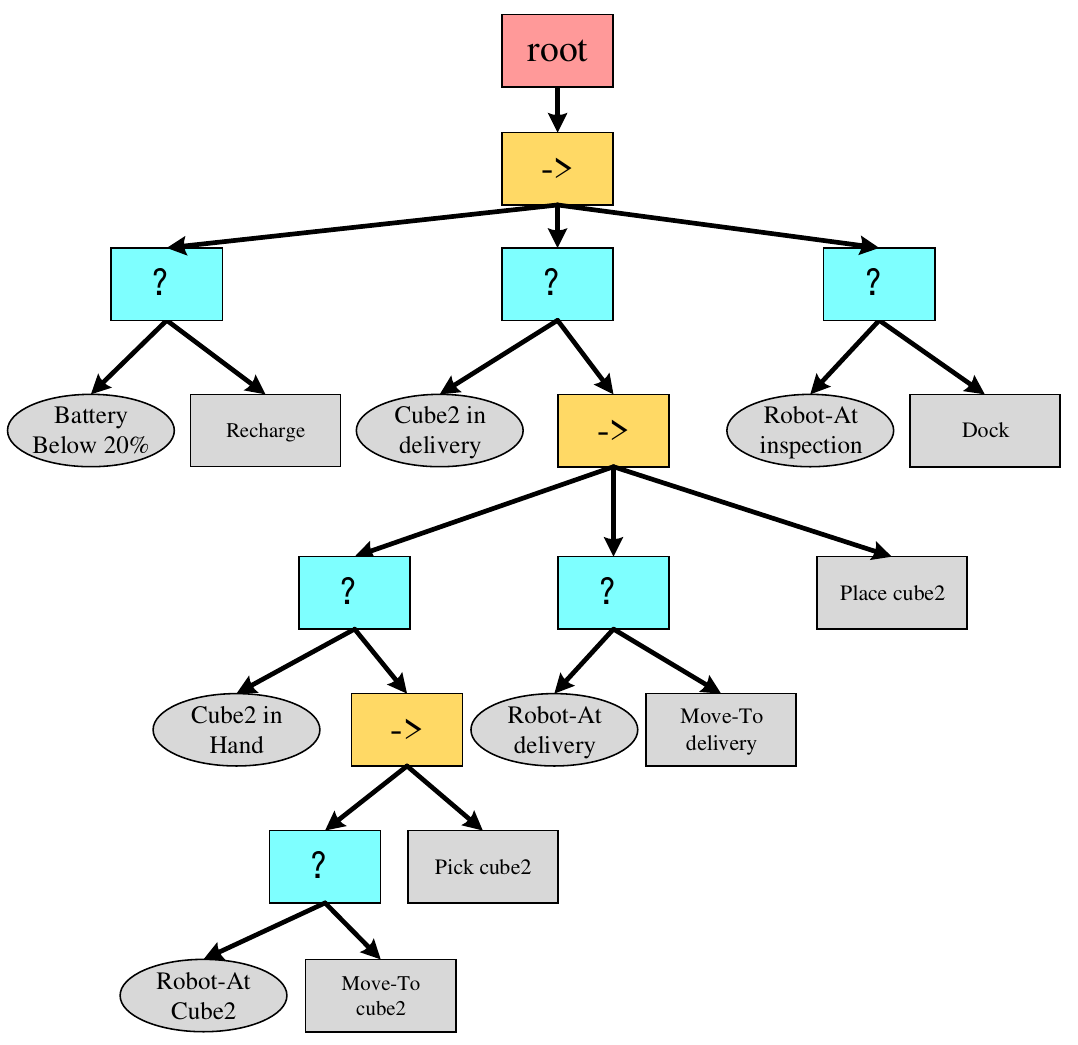} 
    \caption{BT behavioral model of robot.} 
    \label{fig:fig10} 
\end{figure}

To evaluate the conversion quality of LLMDUCF, we conducted simulation experiments under Scenario 2 to compare the performance of different conversion methods. The initial behavior tree employed in the experiments reuses the FSM behavior model of the ROS[8] platform, as shown in Fig. 14.

\begin{table}[h]
    \centering
    \caption{BT to FSM Behavioral Consistency}
    \label{tab:fsm_conversion}
    \begin{tabular}{l c c c c}
        \toprule
        \textbf{} & \textbf{LLMDUCF} & \textbf{NLLM} & \textbf{BSD} & \textbf{CHDS-BT} \\
        \midrule
        Reachability & \checkmark &\checkmark & \checkmark & \checkmark \\
        Completeness  & \checkmark &\text{\sffamily x} & \checkmark & \checkmark\\
        No Additional Behavior & \checkmark &\text{\sffamily x} & \checkmark & \checkmark \\
        Deadlock Absence & \checkmark &\checkmark & \checkmark & \checkmark \\
        Termination Consistency & \checkmark &\text{\sffamily x} & \checkmark & \checkmark \\
        \bottomrule
    \end{tabular}
\end{table}

Behavioral Consistency: Figs. 15–17 show the FSMs generated from BT in Fig. 14 by LLMDUCF, BSD/CHDS-BT, and NLLM, respectively. NuSMV was used to verify reachability, completeness, absence of additional behavior, deadlock freedom, and termination consistency. As shown in Table III, LLMDUCF, BSD, and CHDS-BT satisfy all criteria, indicating behavioral equivalence with the original BT. This was further confirmed in Gazebo, where the original BT and converted FSM produced identical robot trajectories. In contrast, NLLM fails in completeness, absence of additional behavior, and termination consistency, showing that it misses expected paths, introduces redundant transitions, and cannot reliably preserve termination behavior due to insufficient reasoning over deeply nested BT semantics.

\begin{figure}[h] 
    \centering \includegraphics[width=\linewidth]{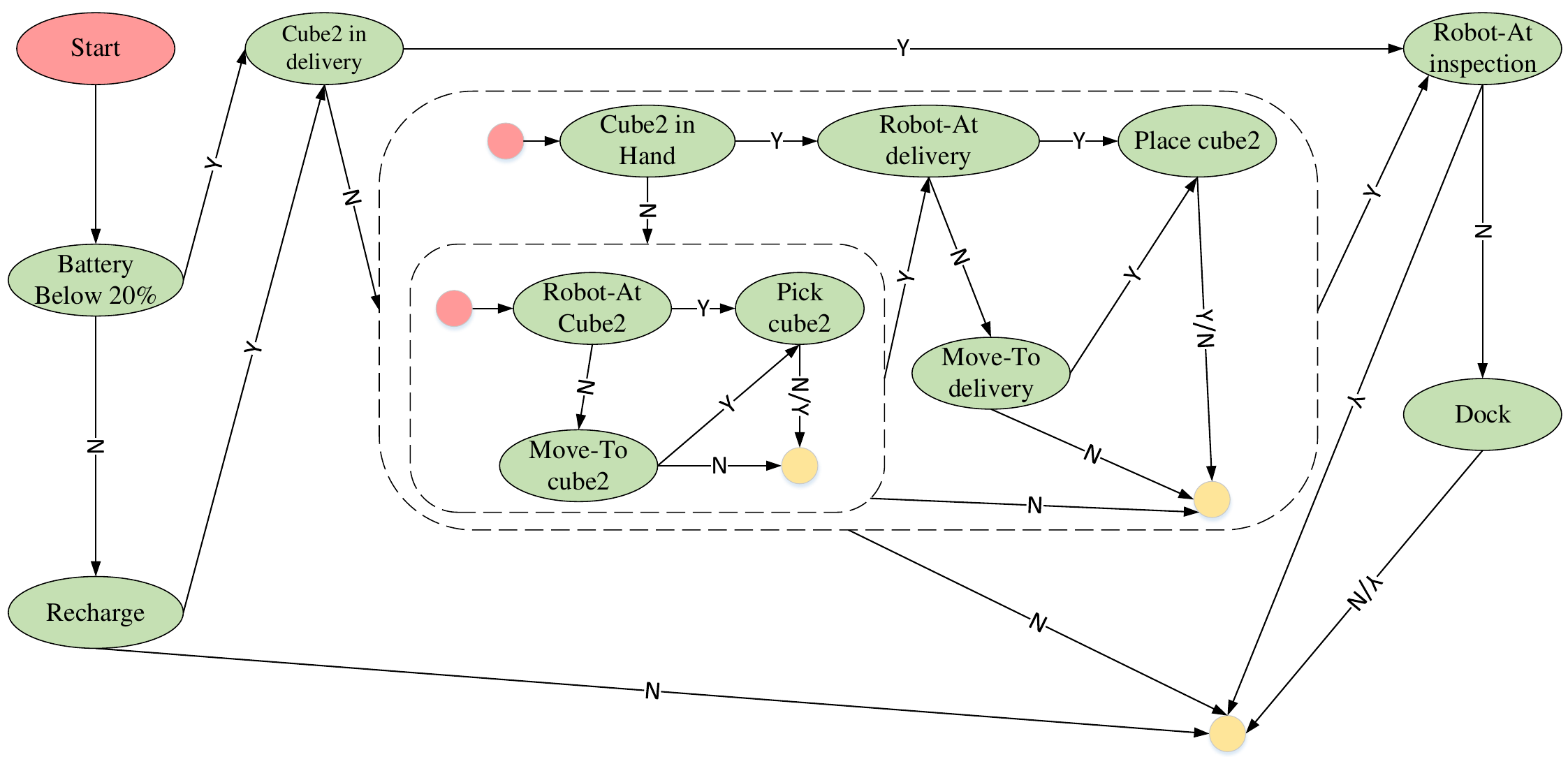} 
    \caption{FSM generated by proposed method.} 
    \label{fig:fig16} 
\end{figure}

\begin{figure}[h] 
    \centering 
    \includegraphics[width=6cm]{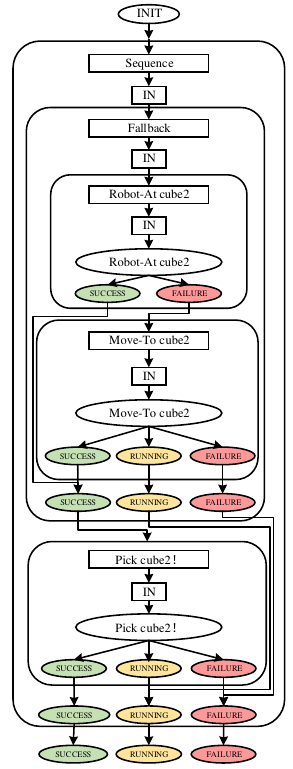} 
    \caption{FSM generated by BSD/CHDS-BT.} 
    \label{fig:fig17} 
\end{figure}

\begin{figure}[h] 
    \centering \includegraphics[width=\columnwidth]{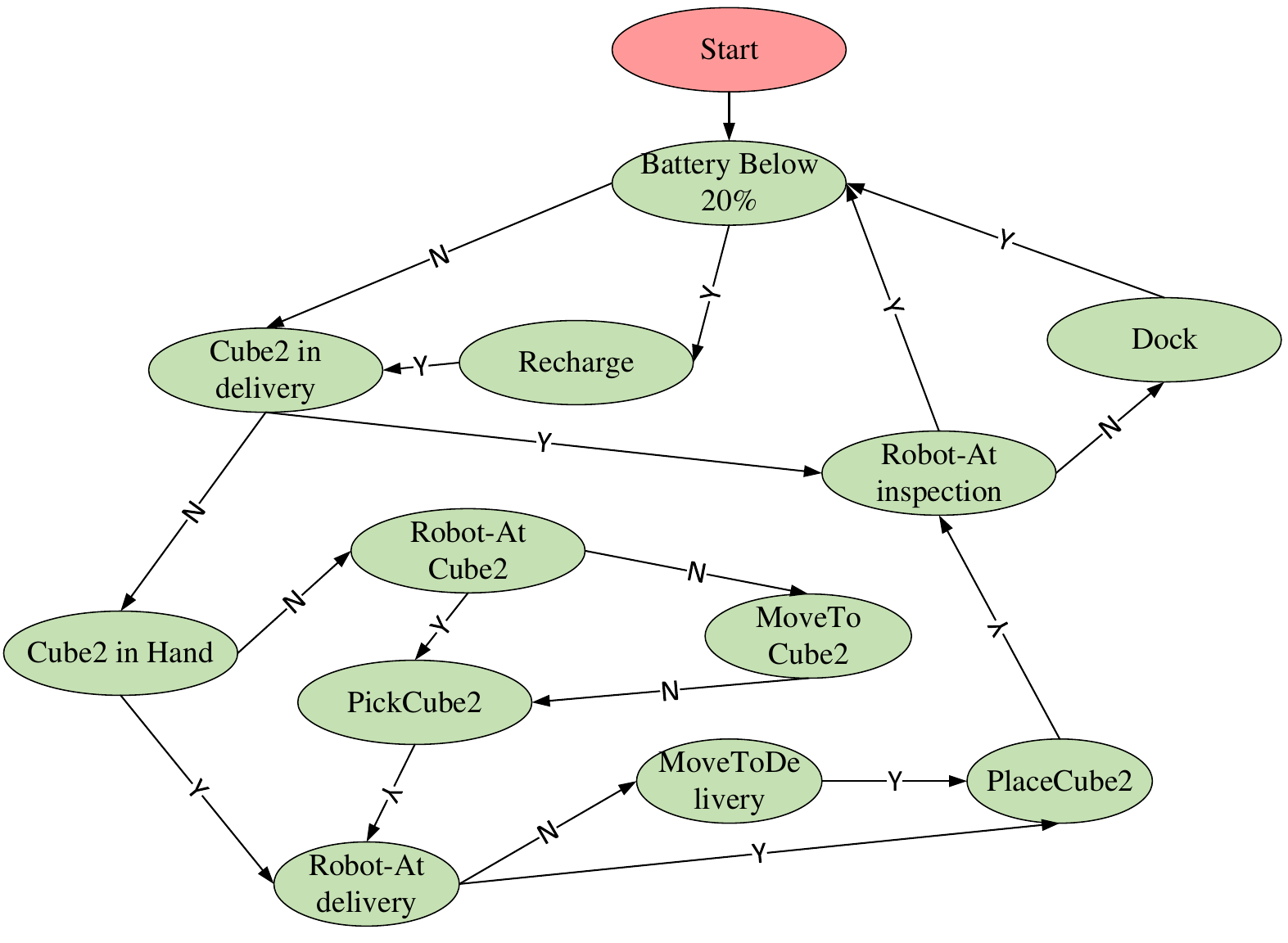} 
    \caption{FSM generated by NLLM.} 
    \label{fig:fig18} 
\end{figure}

\begin{table}[h]
    \centering
    \caption{Comparison of Structural Complexity}
    \label{tab:structural_complexity_iv}
    \begin{tabular}{l c c c}
        \toprule
        & \textbf{LLMDUCF} & \textbf{NLLM} & \textbf{BSD/CHDS-BT} \\
        \midrule
        \makecell{Computational\\ Complexity} & $\text{O}(1)$ & $\text{O}(\textit{n})$ & $\text{O}(1)$ \\
        \addlinespace
        \makecell{Graphical\\ Elements} & $4\text{\textit{M}+2\textit{k}+1}$ & / & $\approx 36\textit{M}$ \\
        \addlinespace
        \makecell{Active\\ Elements} & $4\text{\textit{M}+2\textit{k}+1}$ & / & $\approx 29\textit{M}$ \\
        \bottomrule
    \end{tabular}
\end{table}

Modularity: When adding or removing a state in a standard FSM generated by the NLLM method, consistency between state outputs and transitions must be verified, potentially requiring processing of every state transition with O(\textit{n}) complexity. For the Proposed and converted BSD/CHDS–BT approaches to HFSMs, adding a state incurs the same complexity as in BT. Once the insertion point is determined, the operation is performed only within the current sub-FSM, achieving O(1) time complexity.

\begin{table}[h]
    \centering
    \caption{Edit Distance of FSM}
    \label{tab:edit_distance_fsm}
    \begin{tabular}{l c c c}
        \toprule
        & \textbf{LLMDUCF} & \textbf{NLLM} & \textbf{BSD/CHDS-BT} \\
        \midrule
        \makecell{Tuck Arm\\ subtree} & 4 & 4 & 12 \\
        \addlinespace
        \makecell{Safe-Move-\\To behavior} & 4 & 4 & 4 \\
        \addlinespace
        \makecell{Dock\\ subtree} & 4 & 4 & 17 \\
        \bottomrule
    \end{tabular}
\end{table}

The model scalability of the FSMs was validated through edit distance analysis. When adding Safe-Move-To behavior, Tuck Arm subtree, and Dock subtree behaviors to the FSM, both the Proposed and Baseline methods yielded an edit distance of 4, as illustrated in the structural diagram below. Adding states to the FSM only requires splitting two existing states and connecting the new state. In contrast, the edit distances for the FSMs generated by BSD/CHDS–BT were 4, 12, and 17, respectively. An action node is a subgraph with 1 vertex and 3 edges. The vertex being the node itself and the edges being the 3 transitions, one from each return status to the next child or the parent return statuses. A condition node is similar to an action node, with one edge less because conditions do not return RUNNING. A control node contributes with 1 vertex and 4 edges. The additional edge is the transition from the node to the first child. LLMDUCF exhibits shorter edit distances, greater model modularity, and greater scalability.

Readability: In the FSM generated by the LLMDUCF conversion, each state has two transition paths leading to subsequent states: a successful transition and a failed terminal state. To achieve full reactivity, each state must possess one active self-transition path (i.e., a transition where the state loops back to itself). Under this design, if a finite state machine contains \textit{M} states, its total number of nodes is \textit{M}. The transition paths in the FSM then comprise the following types: 1. \textit{M} self-transitions; 2. \textit{M} success transitions; 3. \textit{M} failure transitions; 4. 2\textit{k} start/end states in sub-FSM; 5. 1 final state. Thus, the total number of graphical elements in the FSM is 4\textit{M}+2\textit{k}+1.

In the FSM generated by the BSD/CHDS–BT conversion, each action corresponds to one conditional node and one back-to-start node, while the number of sequence nodes is approximately half that of the action nodes. This implies that for \textit{M} actions, there will be \textit{M} conditional nodes, \textit{M} back-to-start nodes, and 0.5\textit{M} sequence nodes. Each action node contributes 10 elements, each conditional node contributes 7 elements, and each control node contributes 8 elements, resulting in approximately 29\textit{M} active elements. For calculating the number of graphical elements, two additional elements must be added per node to account for entering the IN state and a corresponding transition path. The total number of graphical elements is 36\textit{M}. The FSM generated by the LLMDUCF conversion offers improved readability. 

In addition, we invited 10 behavioral modeling engineers to evaluate converted behavioral models for cognitive load (scored 1–5, where lower scores indicate reduced cognitive load) and model comprehension time. The experimental results revealed that the LLMDUCF-generated model achieved an average cognitive load score of $1.8{ _{\pm0.3}}$, compared to $4.3{ _{\pm0.6}}$ for traditional methods. The understanding time was reduced by 42$\%$ compared to conventional approaches, validating its superior readability.

\subsection{Performance of FSM-to-BT Conversion}

\begin{figure}[h] 
    \centering \includegraphics[width=\columnwidth]{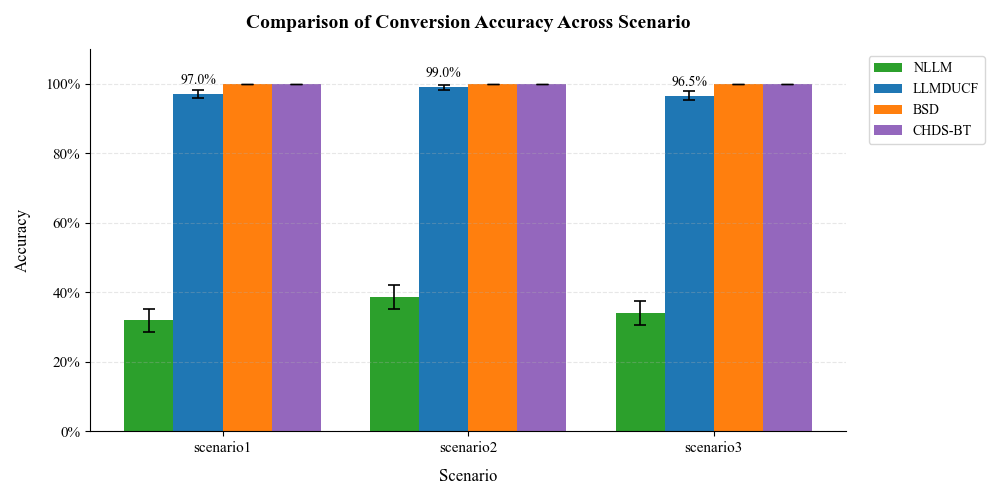} 
    \caption{Comparison of Conversion Accuracy Across Scenario.} 
    \label{fig:fig8} 
\end{figure}

In all evaluation scenarios, traditional conversion methods BSD and CHDS-BT achieved conversion accuracy 100$\%$, attributable to their deterministic mapping mechanisms based on fixed templates, as shown in Fig. 18. The LLMDUCF, when implemented with ChatGPT, achieved significantly higher accuracy than the NLLM in all three scenarios (Scenario 1: 96.95$\%{ _{\pm3.22}}$, Scenario 2: 98.35$\%{ _{\pm1.77}}$, Scenario 3: 95.50$\%{ _{\pm4.37}}$). This performance improvement stems primarily from two key design elements: firstly, we rigorously formalized the typical structure of finite state machines, enabling large language models to clearly and consistently recognize various structural types within input models; secondly, our proposed LEBT templates possess comprehensive expressive power, precisely capturing and converting diverse structures within finite state machines, thereby preserving semantic correctness throughout the conversion process.

\begin{table}[h]
    \centering
    \caption{Conversion Accuracy with 95$\%$ confidence intervals}
    \label{tab:fsm_conversion}
    \begin{tabular}{l c c c}
        \toprule
        \textbf{} & \textbf{scenario1} & \textbf{scenario2} & \textbf{scenario3} \\
        \midrule
        ChatGPT  & $96.95{ _{\pm3.22}}$ &$98.35{ _{\pm1.77}}$ & $95.50{ _{\pm4.37}}$  \\
        DeepSeek &$96.10{ _{\pm3.10}}$ &$97.95{ _{\pm2.82}}$ & $94.05{ _{\pm4.61}}$ \\
        Qwen  &$96.30{ _{\pm3.93}}$ &$96.50{ _{\pm3.09}}$ & $95.70{ _{\pm3.26}}$ \\
        ChatGPT (without SCP) &$31.20{ _{\pm8.94}}$ &$38.70{ _{\pm9.15}}$ & $33.20{ _{\pm9.35}}$  \\
        DeepSeek (without SCP) &$31.35{ _{\pm9.33}}$ &$32.20{ _{\pm9.55}}$ & $33.90{ _{\pm10.65}}$  \\
        Qwen (without SCP) &$32.70{ _{\pm9.10}}$ &$35.60{ _{\pm9.24}}$ & $32.10{ _{\pm9.30}}$  \\
        \bottomrule
    \end{tabular}
\end{table}

\begin{figure}[h] 
    \centering \includegraphics[width=\columnwidth]{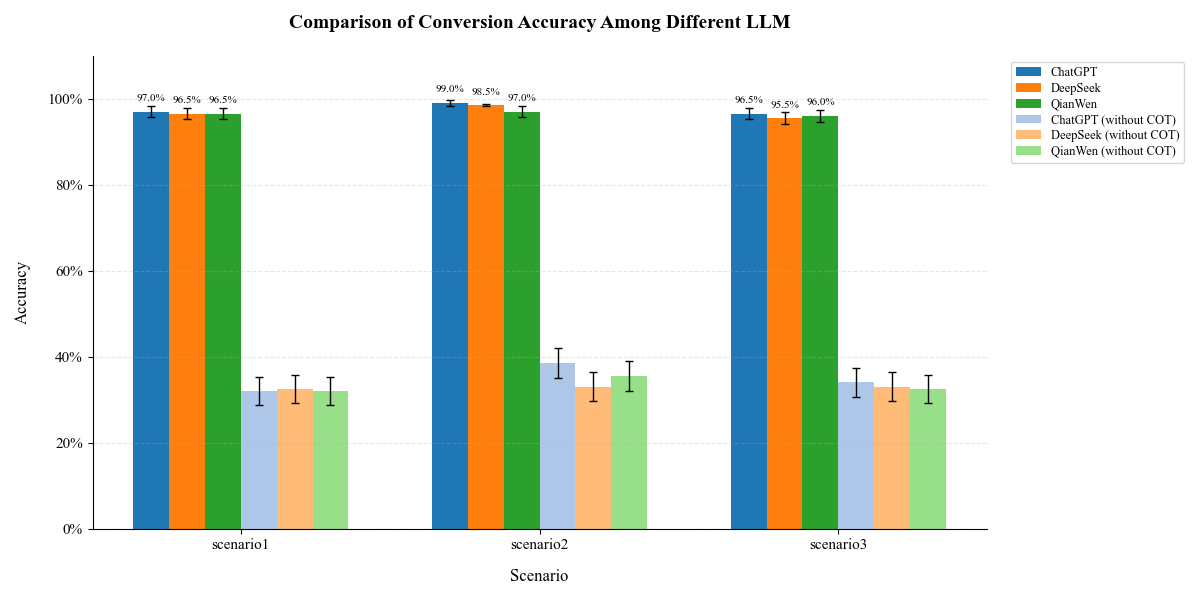} 
    \caption{Comparison of Conversion Accuracy Among Different LLM.} 
    \label{fig:fig8} 
\end{figure}

The accuracy rates achieved by different LLMs in converting FSM to BT across various scenarios are shown in Fig. 19. Across all scenarios, LLM employing SCP prompts consistently outperformed baseline models without SCP. This confirms that SCP prompts specifically designed for conversion tasks effectively guide LLM in understanding semantics and enhancing conversion accuracy. Mainstream LLMs employing SCP demonstrate consistently high performance, with accuracy rates persistently exceeding 94$\%$ and minimal fluctuation. This indicates that the prompt framework possesses strong generalizability and maintains stable output quality when adapted to different models.

\begin{table}[h]
    \centering
    \caption{Ablation Experiments}
    \label{tab:fsm_conversion}
    \begin{tabular}{l c c c c}
        \toprule
        \textbf{} & \textbf{scenario1} & \textbf{scenario2} & \textbf{scenario3} &  \\
        \midrule
        Base & $31.20{ _{\pm8.94}}$ &$38.70{ _{\pm9.15}}$ & $33.20{ _{\pm9.35}}$   \\
        Base+M1 & $33.80{ _{\pm9.74}}$ &$38.65{ _{\pm10.03}}$ & $36.05{ _{\pm10.96}}$ \\
        Base+M2 & $34.30{ _{\pm9.37}}$ &$46.20{ _{\pm10.86}}$ & $41.55{ _{\pm9.69}}$  \\
        Base+M1+M2 & $96.95{ _{\pm3.22}}$ &$98.35{ _{\pm1.77}}$ & $95.50{ _{\pm4.37}}$  \\
        \bottomrule
    \end{tabular}
\end{table}

To elucidate the independent and synergistic effects of each
model within the SCP framework of the LLMDUCF method, we conducted systematic ablation experiments. 
The conversion accuracy results are summarized in Table VII. M1 involved the recognition of the FSM structure, while M2 addressed the definition and mapping relationships of LEBT. The experiments demonstrate that adding only FSM structure recognition yields a limited performance improvement , demonstrating that pure structure recognition alone is insufficient to bridge the semantic conversion gap. Adding LEBT structure mapping alone produces a slightly greater improvement, yet overall accuracy remains below 50$\%$, reflecting incomplete input comprehension. When both components are integrated, the conversion accuracy in all scenarios increases to high levels of 95.50$\%$ to 98.35$\%$. This indicates complementary roles: M1 provides precise FSM input parsing, while M2 enables accurate LEBT structural mapping. Their integration enhances the conversion performance of LLMDUCF. In summary, LLMDUCF relies on the synergistic interaction between the FSM structural recognition and LEBT structural mapping components, rather than the contribution of any single component. Both are indispensable elements for achieving high-precision FSM to BT conversion.

\begin{figure}[h] 
    \centering \includegraphics[width=\columnwidth]{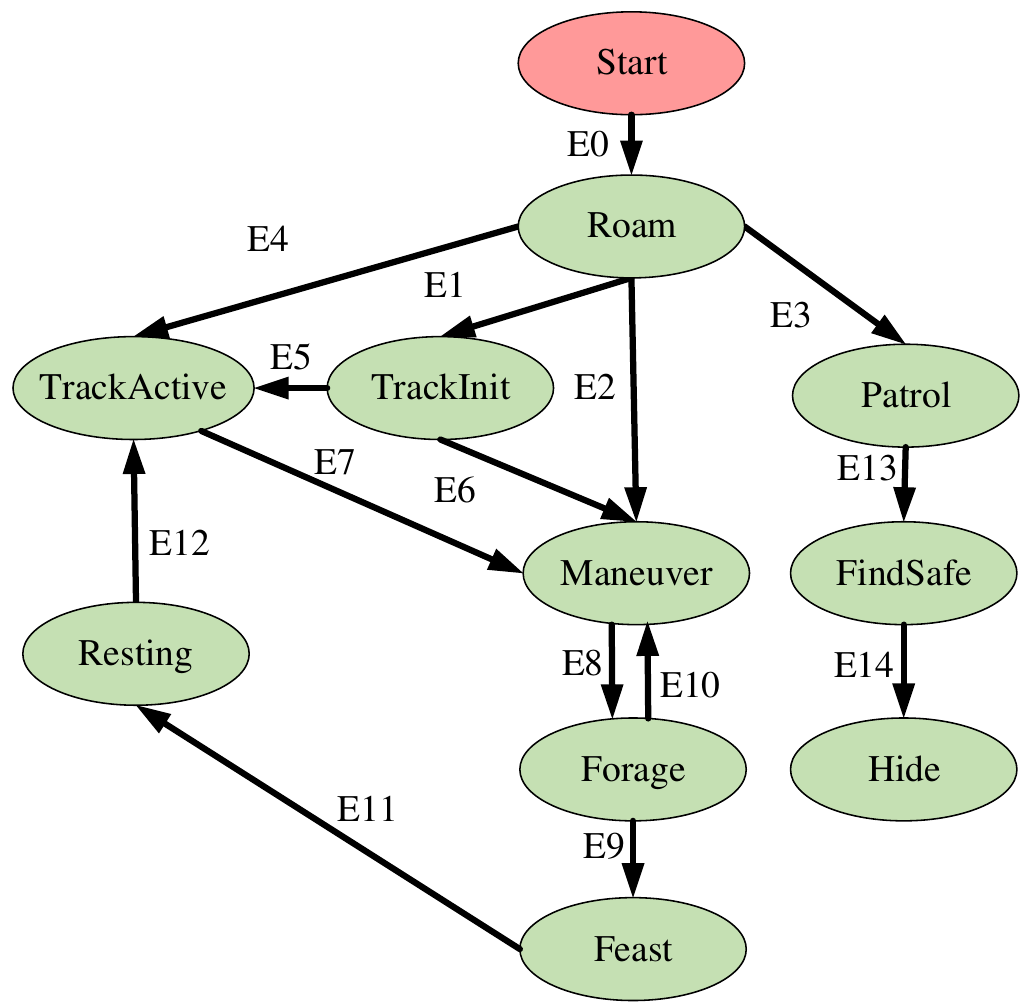} 
    \caption{FSM behavioral model of predator.} 
    \label{fig:fig9} 
\end{figure}

\begin{figure}[!t] 
    \centering \includegraphics[width=\linewidth]
    {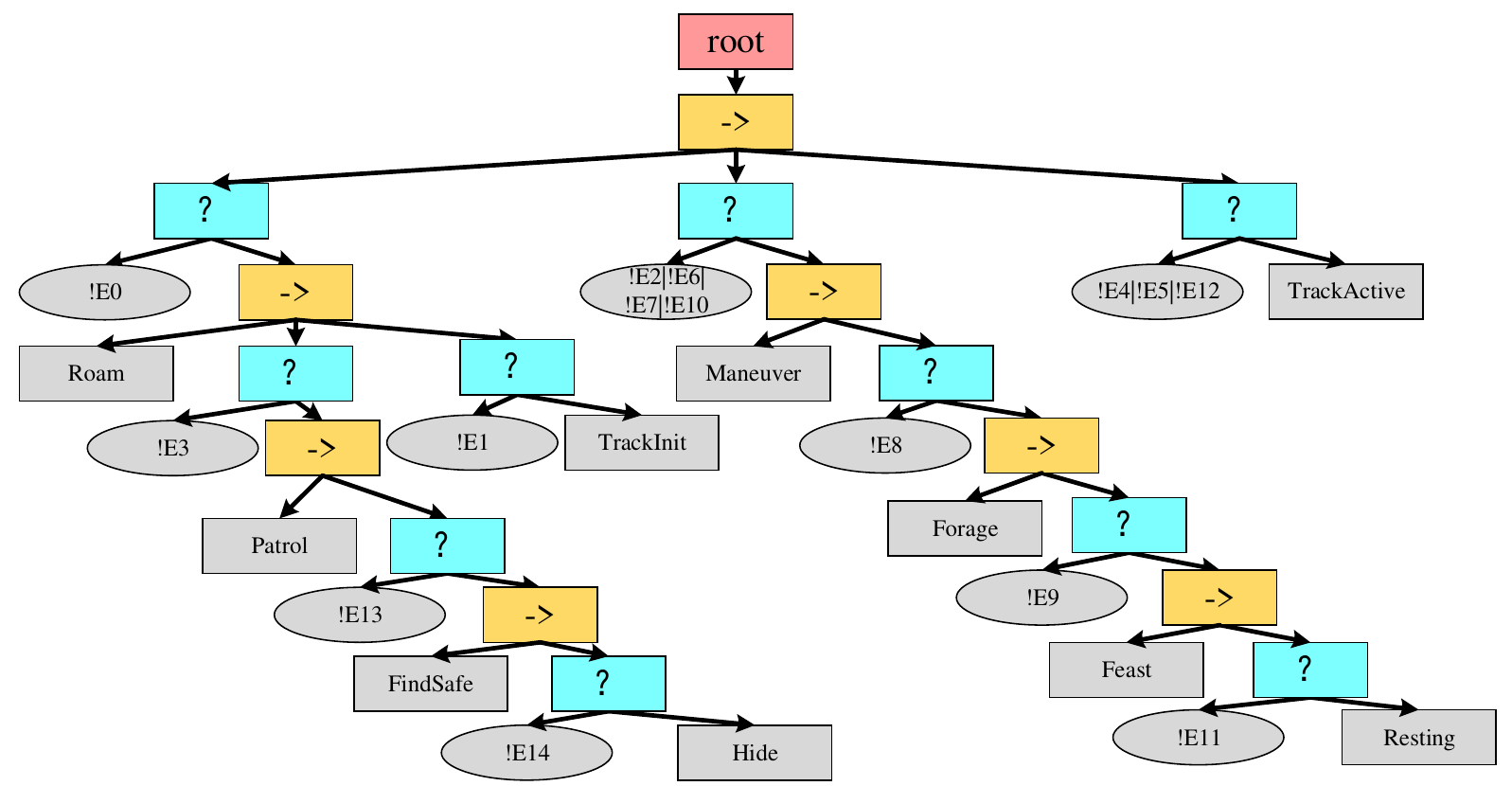} 
    \caption{BT model generated using proposed method.} 
    \label{fig:fig11} 
\end{figure}

To evaluate the conversion quality of LLMDUCF, we conducted simulation experiments under Scenario 3 to compare the performance of different conversion methods. The initial FSM employed in the experiments was designed manually , as shown in Fig. 20.

\begin{table}[h]
    \centering
    \caption{FSM to BT Behavioral Consistency}
    \label{tab:fsm_conversion}
    \begin{tabular}{l c c c c}
        \toprule
        \textbf{} & \textbf{LLMDUCF} & \textbf{NLLM} & \textbf{BSD} & \textbf{CHDSs-BT} \\
        \midrule
        Reachability & \checkmark &\text{\sffamily x} & \checkmark & \checkmark \\
        Completeness  & \checkmark &\text{\sffamily x} & \checkmark & \checkmark\\
        No Additional Behavior & \checkmark &\text{\sffamily x} & \checkmark & \checkmark \\
        Deadlock Absence & \checkmark &\checkmark & \checkmark & \checkmark \\
        Termination Consistency & \checkmark &\text{\sffamily x} & \checkmark & \checkmark \\
        \bottomrule
    \end{tabular}
\end{table}

Behavioral Consistency: Figs. 21–24 show the BTs generated from the FSM in Fig. 20 by LLMDUCF, BSD, CHDS-BT and NLLM, respectively. As shown in Table VIII, NuSMV verification confirms that LLMDUCF, BSD, and CHDS-BT satisfy all behavioral consistency criteria and are equivalent to the original FSM. The predator-prey simulation further shows consistent trajectory behavior across these models. In contrast, NLLM fails in reachability, completeness, absence of additional behavior, and termination consistency, indicating semantic loss in FSM-to-BT conversion, particularly for loop structures. LLMDUCF preserves these behaviors by guiding the LLM to identify loop, chain, and branch structures and map them into the proposed LEBT representation.

\begin{figure}[h] 
    \centering \includegraphics[width=\columnwidth]{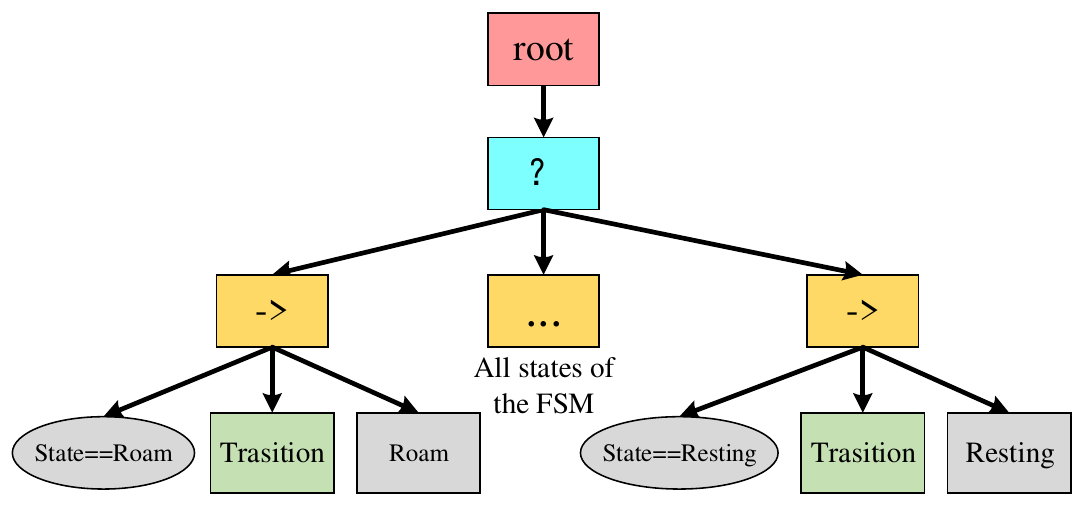} 
    \caption{BT generated by BSD.} 
    \label{fig:fig12} 
\end{figure}

\begin{figure}[h] 
    \centering \includegraphics[width=\columnwidth]{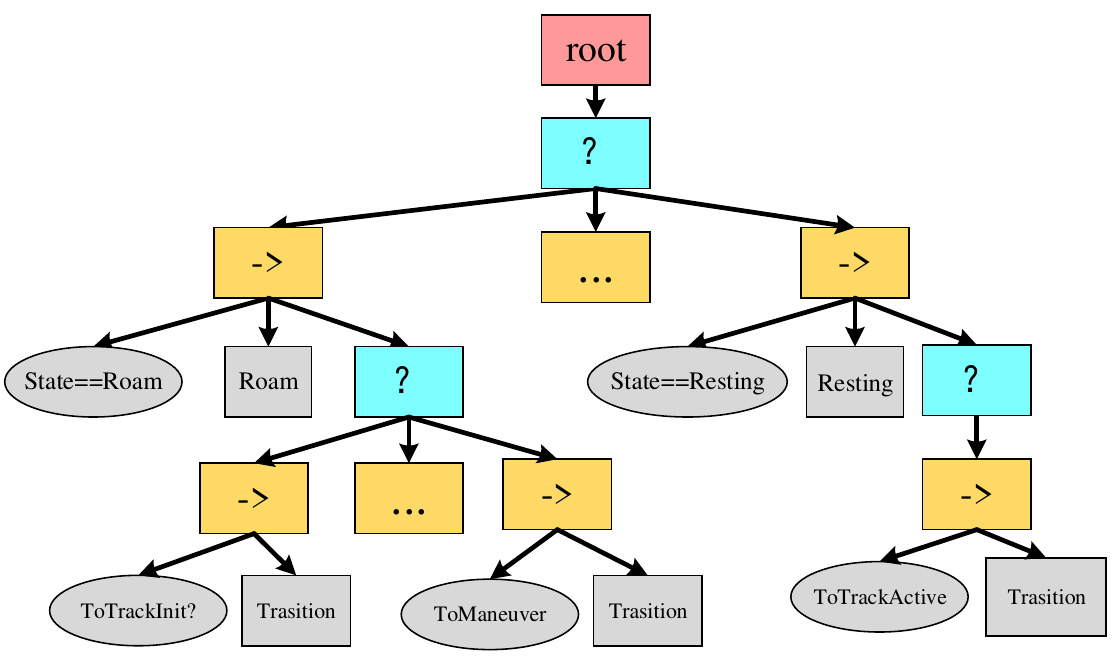} 
    \caption{BT generated by CHDS–BT.} 
    \label{fig:fig13} 
\end{figure}

\begin{figure}[h] 
    \centering \includegraphics[width=\columnwidth]{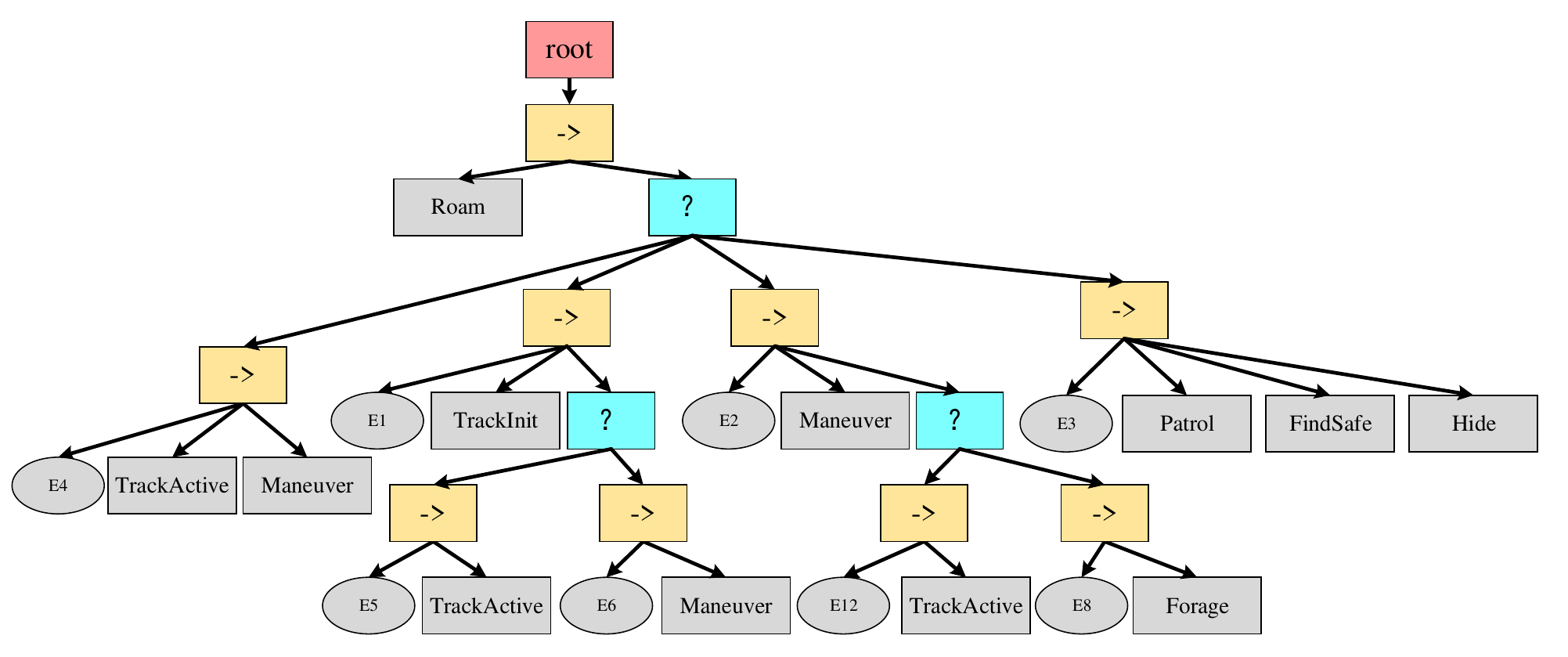} 
    \caption{BT generated by NLLM.} 
    \label{fig:fig14} 
\end{figure}

\begin{table}[h]
    \centering
    \caption{Comparison of Structural Complexity}
    \label{tab:structural_complexity}
    \begin{tabular}{l c c c c}
        \toprule
        & \textbf{LLMDUCF} & \textbf{NLLM} & \textbf{BSD} & \textbf{CHDS-BT} \\
        \midrule
        \makecell{Computational\\ Complexity} & $\text{O}(1)$ & $\text{O}(1)$ & $\text{O}(n)$ & $\text{O}(n^2)$ \\
        \addlinespace
        \makecell{Graphical\\ Elements} & $\approx 7\textit{M-1}$ & / & $8\textit{M+1}$ & $8\textit{M+1+6k}$ \\
        \addlinespace
        \makecell{Active\\ Elements} & $\approx 3.5\textit{M}$ & / & $4\textit{M+1}$ & $4\textit{M+1+3k}$ \\
        \bottomrule
    \end{tabular}
\end{table}
Modularity: In BT, inserting a new node requires adding it to the child node list of its designated parent control node. The BT generated by the LLMDUCF and NLLM methods maintains clear decoupling between the child nodes, allowing insertion and deletion operations to be performed by directly accessing the parent node. This yields an optimal time complexity of O(1) for both operations.In contrast, BT produced by the BSD method incorporates transition relationships between nodes, where the next executable node is determined through explicit transition nodes. Inserting a new node necessitates traversing all transition nodes to update path connections, leading to a time complexity of O(\textit{n}), in addition to requiring modifications to the transition node source code.
The CHDS-BTs also maintain node transition links. However, in this case, it requires traversing all nodes and their associated transition paths to establish new connections, resulting in a time complexity of O(\textit{n}²). The higher complexity underscores the maintenance challenges for dynamic and scalable autonomous systems, where frequent model updates are often required.
\begin{table}[h]
    \centering
    \caption{Edit Distance of BT}
    \label{tab:edit_distance_bt}
    \begin{tabular}{l c c c c}
        \toprule
        & \textbf{LLMDUCF} & \textbf{NLLM} & \textbf{BSD} & \textbf{CHDS-BT} \\
        \midrule
        \makecell{Fire\\ subtree} & 6 & 6 & $8+\textit{k}$ & $8+6\textit{k}$ \\
        \addlinespace
        \makecell{Gliding\\ behavior} & 2 & 2 & 2 & 2 \\
        \addlinespace
        \makecell{Loading\\ subtree} & 8 & 8 & $10+\textit{k}$ & $10+6\textit{k}$ \\
        \bottomrule
    \end{tabular}
\end{table}

\begin{figure}[ht] 
    \centering \includegraphics[width=\linewidth]{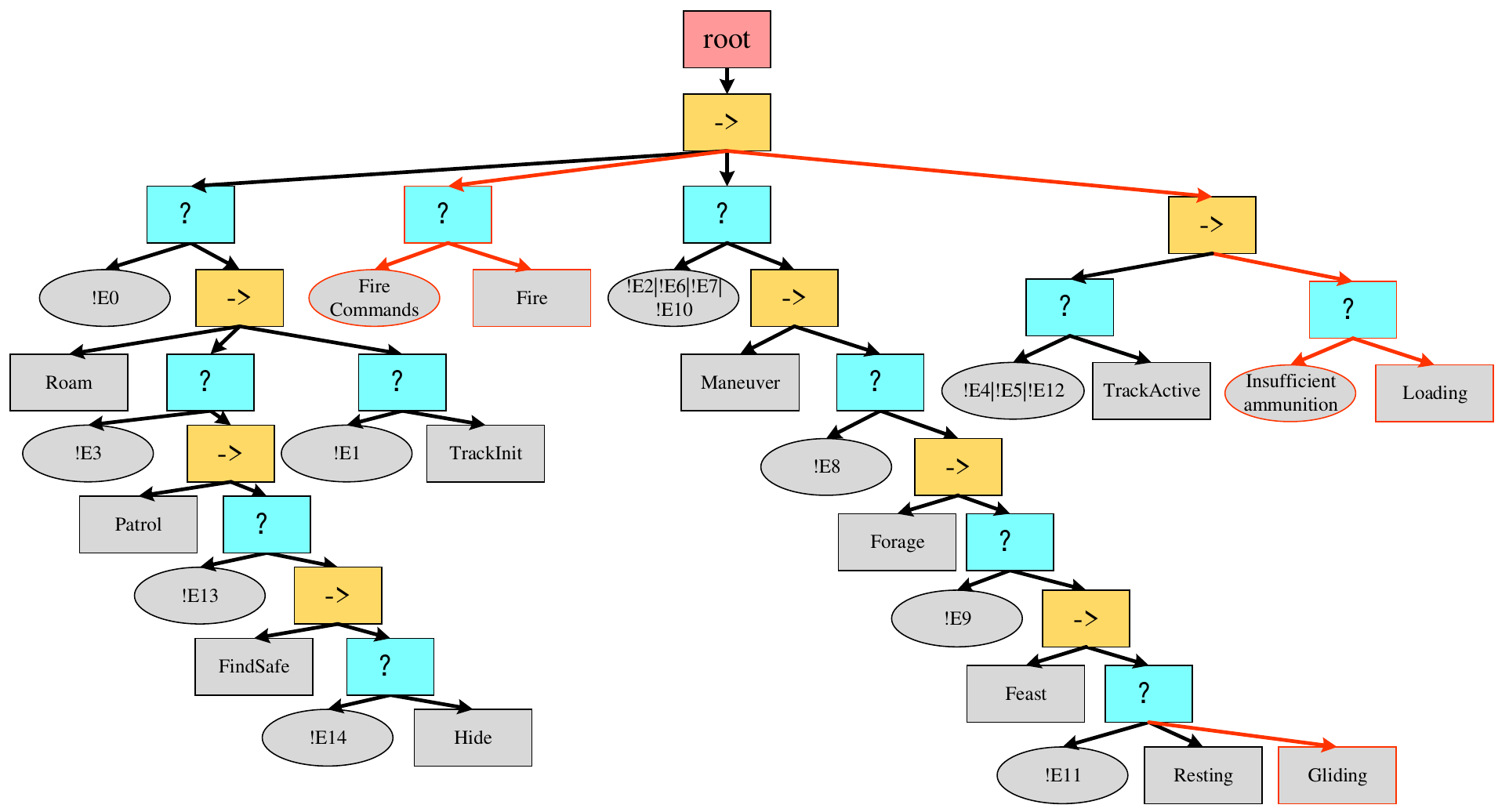} 
    \caption{Different cases of node addition in a BT. The added nodes are highlighted in red.} 
    \label{fig:fig15} 
\end{figure}
The edit distance serves as a key metric for evaluating the scalability and maintainability of BT models when extending their functionality. As illustrated in Fig. 25, when introducing new behaviors—namely the Gliding behavior, the Fire sub-tree, and Loading sub-tree behaviors into the base BT, the LLMDUCF and NLLM methods yield edit distances of 2, 6, and 8, respectively. This low modification cost stems from the node independence in these BT: adding a node or sub-tree only involves inserting the corresponding nodes and edges, without impacting existing structures.
In contrast, the BSD-generated BT produces edit distances of 2, 8+\textit{k}, and 10+\textit{k}, where \textit{k} denotes the number of existing transition relationships associated with the current node. Each action insertion requires adding 4 nodes and 4 edges. More importantly, incorporating a new sub-tree requires creating a new transition node and updating transition linkages across other sub-trees, increasing both structural and coding overhead.
The CHDS-BT method results in even greater edit distances of 2, 8+6\textit{k}, and 10+6\textit{k}. Here, introducing a new sub-tree demands creating an entire conversion sub-tree to reestablish transition connections, significantly increasing operational complexity. These results highlight the advantages of LLMDUCF in supporting efficient and low-overhead model extension.

Readability: To quantitatively assess the readability of generated BT, we introduce a metric based on the number of graphical elements within the model structure. Let \textit{M} denote the number of action nodes. In the proposed LLMDUCF method, each action node is associated with one selector node and one condition node. The number of sequence nodes is variable, but can be empirically estimated to be half the number of action behaviors. Thus, the total number of nodes is approximately 3.5\textit{M}. To account for all graphical components, edges connecting the parent and child nodes must also be included, resulting in 3.5\textit{M}-1 edges. This results in a total of 7\textit{M}-1 graphical elements.
In contrast, the NLLM method produces non-deterministic BT structures, making it impossible to consistently quantify graphical elements. The BSD and CHDS-BT methods yield totals of 8\textit{M}+1 and 8\textit{M}+1+6\textit{k} elements, respectively, where \textit{k} reflects additional transition relations. The LLMDUCF method significantly reduces the graphical complexity compared to these alternatives, directly contributing to better readability.

\subsection{Sensitivity and Robustness Evaluation}

\begin{table}[t]
    \centering
    \caption{Comparison of Version Sensitivity}
    \label{tab:structural_complexity}
    \begin{tabular}{l c c}
        \toprule
        & \textbf{BT to FSM} & \textbf{FSM to BT} \\
        \midrule
        \makecell{GPT(gpt-4-turbo)} & 93.85$\%$ & 96.20$\%$  \\
        \addlinespace
        \makecell{GPT(gpt-4o)} & 94.55$\%$ & 97.05$\%$  \\
        \addlinespace
        \makecell{DeepSeek(v3.1)} & 94.25$\%$ & 95.30$\%$  \\
        \addlinespace
        \makecell{DeepSeek(v3.2)} & 95.10$\%$ & 96.15$\%$  \\
        \addlinespace
        \makecell{Qwen(v2)} & 92.90$\%$ & 95.45$\%$  \\
        \addlinespace
        \makecell{Qwen(v3)} & 93.80$\%$ & 95.75$\%$ \\
        \bottomrule
    \end{tabular}
\end{table}

This section provides a systematic and comprehensive analysis of the proposed LLMDUCF framework, thoroughly examining its practical performance in terms of version sensitivity, edge case handling, and input-output compatibility.

Sensitivity: We conducted comparative experiments on the version sensitivity of large language models, contrasting the latest official stable versions with their preceding iterations, the results are shown in Table XI. Experimental results demonstrate that the proposed LLMDUCF framework exhibits less than 3$\%$ fluctuations in conversion accuracy between different versions of the same model. This deviation falls within the experimentally acceptable error range and does not affect the conclusions of the core research presented here. The low version sensitivity of this framework stems from the dual-constraint mechanism of SCP prompts and the post-processing structural verification layer. Their synergistic action effectively mitigates subtle logical deviations arising from large model version iterations, thereby ensuring the stability and consistency of bidirectional conversion results between FSM and BT across versions.

\begin{table}[h]
    \centering
    \caption{Retry Statistics of Failed Conversion Cases}
    \label{tab:fsm_conversion}
    \begin{tabular}{l c c c}
        \toprule
        \textbf{} & \textbf{Failure Cases} & \textbf{Total Retries} & \textbf{Average Retries} \\
        \midrule
        BT-FSM(ChatGPT) & 27 & 42 & 1.56  \\
        BT-FSM(DeepSeek)  & 31 & 69 & 2.23 \\
        BT-FSM(Qwen) & 35 & 76 & 2.17 \\
        FSM-BT(ChatGPT) & 18 & 41 & 2.28 \\
        FSM-BT(DeepSeek) & 24 & 53 & 2.21 \\
        FSM-BT(Qwen) & 23 & 51 & 2.22 \\
        \bottomrule
    \end{tabular}
\end{table}

Failure Cases and Recovery Methods: Table XII reports the number of first-pass failures, total retry attempts, and average retries per failed case in the three experimental scenarios. For both BT-to-FSM and FSM-to-BT conversion, the average number of retries ranges from 1.56 to 2.28, showing that most failed initial conversions can be corrected with a small number of feedback-guided regeneration attempts, achieving 100\% accuracy. Behavioral model conversion is primarily an offline engineering activity performed during system design, migration, and refactoring. Its objective is to obtain a formally verified correct model, rather than to satisfy runtime latency constraints. Consequently, NuSMV-based verification, error feedback, and regeneration constitute an acceptable processing workflow. 

Conversion failures in LLMDUCF primarily stem from hallucinations in LLM. Although this paper employs SCP prompts and conversion constraints, in a small number of complex or edge cases, the LLM may still fail to strictly follow certain prompts, leading to errors in structural mapping or the identification of conversion units. In the conversion from FSM-to-BT, a common error is the presence of redundant structures in the generated LEBT. For example, when a particular state is identified as both an entry state and a loop state, the LLM may generate duplicate sub-structures for the same state, leading to logical errors. This issue can be resolved by incorporating a node uniqueness constraint into the prompt, namely requiring that each state correspond to only one node in the LEBT, and that duplicate nodes be merged. In BT-to-FSM conversion, the primary error is the incorrect identification of transition units. LLMs sometimes treat every control node as an independent transition unit, resulting in the generation of too many sub-FSMs and an overly complex structure. This issue can be remedied by refining the definition of transition units and providing reference examples to guide the LLM in correctly partitioning transition units according to hierarchical rules. Therefore, although LLM outputs may still exhibit instability in a small number of complex structures, these errors can usually be corrected through targeted prompt optimization and enhanced structural constraints.

\begin{table}[htbp]
    \centering
    \caption{Input Representation Generalizability}
    \label{tab:io_robustness}
    \begin{tabular}{l c c c c}
        \toprule
        & \textbf{LLMDUCF} & \textbf{NLLM} & \textbf{BSD} & \textbf{CHDS-BT} \\
        \midrule
        Text File & \checkmark & \checkmark & \checkmark & \checkmark \\
        \addlinespace
        Image & \checkmark & \checkmark & \text{\sffamily x} & \text{\sffamily x} \\
        \addlinespace
        Table & \checkmark & \checkmark & \text{\sffamily x} & \text{\sffamily x} \\
        \addlinespace
        Unseen Encoding & \checkmark & \checkmark & \text{\sffamily x} & \text{\sffamily x} \\
        \addlinespace
        New Conventions & \checkmark & \checkmark & \text{\sffamily x} & \text{\sffamily x} \\
        \bottomrule
    \end{tabular}
\end{table}

Input Representation Generalizability: LLMDUCF and NLLM are capable of processing three types of input representations—text files, images, and structured tables—and can adapt to unseen encoding schemes and new modeling conventions, demonstrating that methods based on large language models possess strong generalization capabilities regarding input representations. In contrast, BSD and CHDS-BT only support predefined text-based structured inputs; when inputs are presented as images, tables, unseen encodings, or new expression conventions, neither can perform the conversion directly and typically require additional format parsing, model preprocessing, or manually extended conversion rules. LLMDUCF can further improve the behavioral correctness and stability of conversion results across different input representations, thereby enhancing the flexibility of the model conversion process and interoperability between tools.

\subsection{Experiment Summary and Discussion}
The multi-scenario experimental evaluation comprehensively validates the performance and intelligent characteristics of the proposed LLMDUCF in bidirectional FSM-BT conversion. The results demonstrate a clear trade-off profile between traditional methods, NLLM, and our framework, highlighting its distinct value proposition.

Performance and Generality: While traditional rule-based methods (BSD, CHDS-BT) achieve 100$\%$ conversion accuracy through handcrafted deterministic templates, they inherently lack generalizability and require expert knowledge for each new model or scenario. In contrast, NLLM approaches offer automation but suffer from low and unstable accuracy due to insufficient comprehension of complex behavioral logic. The LLMDUCF strikes an effective balance: it achieves high conversion accuracy (ranging from 90.65$\%$ to 98.35$\%$ across all tasks) by leveraging the LLM's parsing capability while constraining its reasoning with domain-specific rules and structural strategies. This enables robust performance across diverse scenarios—from basic symbols to embodied robotic tasks and dynamic game AI—without the need for scenario-specific template engineering. Furthermore, the framework offers advantages such as strong version stability, adaptability to complex structures, and compatibility with multimodal input and output. Although there are minor fluctuations in accuracy in a small number of edge cases, it demonstrates high overall reliability and practical value.

Conversion Quality and Model Utility: Beyond accuracy, the quality of the generated models is paramount for practical application. In terms of behavioral consistency, the LLMDUCF-generated models fully satisfy all specified requirements (Tables III$\&$VIII), demonstrating semantic equivalence to the original models. Regarding Modularity, operations such as adding or removing nodes in LLMDUCF-generated models exhibit lower computational complexity (often O(1)) and shorter edit distances compared to models from traditional methods (Tables V$\&$X), indicating a superior ease of modification and extension. For Readability, LLMDUCF-generated models contain significantly fewer graphical and active elements than those produced by traditional methods (Tables IV$\&$IX), leading to clearer visual representations and reduced cognitive load for engineers. Finally, LLMDUCF natively supports multimodal input, which improves flexibility and ease of use compared to traditional methods designed for specific formats.

Cost Analysis: Although LLMDUCF has higher per-conversion latency than traditional rule-based methods, i.e., 60–70 s versus 0.1–0.5 s, behavioral model conversion in consumer-grade AIS is typically a low-frequency engineering task conducted during system design, migration, or refactoring rather than runtime execution. Therefore, overall engineering efficiency is more critical than single-instance latency. For 100 models, traditional methods require approximately 68 h in total, including approximately 48 h to develop a dedicated converter and 20 h to preprocess input-format, with an estimated cost of $\$$430–440. In contrast, LLMDUCF does not require dedicated converter development or strict preprocessing, completing the same workload in about 2 h with an API cost of approximately $\$$5–10(taking ChatGPT (gpt-4o) as an example, a single conversion consumes approximately 500 to 1000 tokens, with API call costs of roughly $\$$0.05 to 0.10 per call). Thus, LLMDUCF significantly reduces development time, labor effort, and overall cost, making it well suited for small-scale or occasional conversion of a behavioral model in consumer AIS scenarios. Furthermore, in the development of consumer-grade AIS, behavioral model conversion is primarily applied during the system design, model migration, and model refactoring phases. Once conversion is complete, the target behavior model can be used directly or in a further way maintained, and no additional model conversion is required during AIS runtime. Therefore, the requirements for real-time performance are relatively lenient.

In addition, we further validated LLMDUCF on consumer-grade hardware using a locally deployed Qwen-35B model via Ollama 3.1 on a PC with an RTX 3060 GPU and 16 GB RAM. In 20 representative behavioral models, local deployment achieved an accuracy of 80$\%$  for FSM-to-LEBT conversion and 85$\%$ for BT-to-FSM conversion, with an average latency of approximately 2 min per model. Although its performance is lower than that of cloud-based large-scale models due to weaker structured reasoning and hardware-constrained quantization/offloading, the results confirm the feasibility of low-cost, cloud-independent deployment for consumer-grade autonomous intelligent systems.

Limitations and Future Prospects: LLMDUCF is only applicable to behavioral models with a finite set of behaviors, complete structures, and logical consistency. For models containing anomalies, ambiguities, or semantic vagueness, such issues must be resolved prior to formal transformation. In addition, the current framework supports only fundamental structural transformations of behavioral models. For more complex model variants, we shall refine and extend our approach in future work: Parallel nodes may be mapped as a set of concurrently executed sub-FSMs, or a runtime environment supporting concurrent semantics may be introduced to achieve synchronous scheduling and state synchronization for multi-branch behaviors;
For decorator nodes (such as UntilSuccess, Repeat, etc.), these may be modeled as loop states within the FSM featuring specific exit conditions, with decorator logic expressed equivalently through customized transition constraints; The core of hierarchical FSM transformation lies in hierarchical nesting and cross-layer transitions. A hierarchical FSM can be decoupled into a set of FSMs in the order of top-level and sub-levels, starting from the lowest-level FSM and transforming it into a sub-tree according to the transformation rules, ultimately constructing a composite LEBT.
The aforementioned extensions necessitate syntactic expansion of the existing LEBT and FSM template alongside the design of more refined LLM prompt engineering. This will guide the model in understanding and processing these advanced behavioral semantics while ensuring semantic consistency and executability of the conversion model.

\section{CONCLUSION}
This paper presents an LLM-driven unified conversion framework that enables a fully automated and semantically consistent bidirectional transformation between FSM and BT. The proposed LEBT structure effectively preserves loop behavior in FSM-to-BT conversion, while a compression and hierarchical rule strategy successfully mitigates the sub-FSM explosion problem in the reverse direction. Comprehensive experiments in multiple autonomous decision-making scenarios demonstrate that LLMDUCF achieves high conversion accuracy (90.65$\%$–98.35$\%$) while generating models with superior structural quality—exhibiting full path coverage, lower modification complexity, and improved readability compared to traditional methods. This work provides a practical and automated solution for enhancing the interoperability of behavioral models in AIS, it provides practical tools for the transfer, verification, and maintenance of autonomous behavior logic in service robots, smart home devices, game agents, and smart electric vehicles.

\end{document}